\RequirePackage{fix-cm}
\documentclass{llncs}
\usepackage{makeidx}  % allows for indexgeneration
\usepackage{graphicx}
\usepackage{hyperref}

\usepackage{url}
\usepackage{xcolor}
\usepackage[utf8]{inputenc}
\usepackage{multirow}
\usepackage[cmex10]{amsmath}
\usepackage[tight,footnotesize]{subfigure}
\usepackage{times}
\usepackage{epsfig}
\usepackage{amssymb}
\usepackage{ifthen}
\usepackage{booktabs}
\usepackage{array}
\usepackage{siunitx}
\usepackage{collcell}
\usepackage{hhline}
\usepackage{pgf}
\usepackage{colortbl}
\usepackage[ruled]{algorithm2e}

\newcommand{\myparagraph}[1]{\textbf{#1} ---}

\definecolor{cwblue1}{rgb}{0.27,0.427,0.623}
\definecolor{cwblue2}{rgb}{0.286,0.454,0.658}
\definecolor{cwblue3}{rgb}{0.733,0.811,0.905}
\def\colorModel{hsb} %You can use rgb or hsb

\newcommand\ColCell[1]{
  \pgfmathparse{#1<50?1:0}  %Threshold for changing the font color into the cells
    \ifnum\pgfmathresult=0\relax\color{white}\fi
  \pgfmathsetmacro\compA{0}      %Component R or H
  \pgfmathsetmacro\compB{#1/100} %Component G or S
  \pgfmathsetmacro\compC{1}      %Component B or B
  \edef\x{\noexpand\centering\noexpand\cellcolor[\colorModel]{\compA,\compB,\compC}}\x #1
  }
\newcolumntype{E}{>{\collectcell\ColCell}m{0.4cm}<{\endcollectcell}}  %Cell width

\newcommand{\ba}{\mbox{\boldmath$a$}}

\newcommand{\bd}{\mbox{\boldmath$d$}}

\newcommand{\bl}{\mbox{\boldmath$l$}}

\newcommand{\bo}{\mbox{\boldmath$o$}}

\newcommand{\bs}{\mbox{\boldmath$s$}}
\newcommand{\bt}{\mbox{\boldmath$t$}}

\newcommand{\bA}{\mbox{\boldmath$A$}}

\newcommand{\bD}{\mbox{\boldmath$D$}}

\begin{document}

\title{Learning 3D Navigation Protocols on Touch Interfaces with Cooperative Multi-Agent Reinforcement Learning%\thanks{Grants or other
% Touch Interaction Design for 3D Navigation with Cooperative Multi-Agent Reinforcement Learning
}
\titlerunning{Learning 3D Navigation Protocols with Cooperative Multi-Agent RL}        % if too long for running head

\author{Quentin Debard\inst{1} \and
      	Jilles Steeve Dibangoye\inst{2} \and
        Stéphane Canu\inst{3} \and
        Christian Wolf\inst{4}
}
\authorrunning{Quentin Debard et al.}

\institute{Itekube, LIRIS,
           \email{quentin.debard@itekube.com}
           \and
           Inria, CITI-Lab, INSA-Lyon,
           \email{jilles-steeve.dibangoye@insa-lyon.fr}
           \and
           LITIS, INSA-Rouen,
           \email{stephane.canu@insa-rouen.fr}
           \and
           LIRIS, Inria, CITI-Lab, INSA-Lyon,
           \email{christian.wolf@insa-lyon.fr}
}

\maketitle

\begin{abstract}
Using touch devices to navigate in virtual 3D environments such as computer assisted design (CAD) models or geographical information systems (GIS) is inherently difficult for humans, as the 3D operations have to be performed by the user on a 2D touch surface. This ill-posed problem is classically solved with a fixed and handcrafted interaction protocol, which must be learned by the user. We propose to automatically learn a new interaction protocol allowing to map a 2D user input to 3D actions in virtual environments using reinforcement learning (RL). A fundamental problem of RL methods is the vast amount of interactions often required, which are difficult to come by when humans are involved. To overcome this limitation, we make use of two collaborative agents. The first agent models the human by learning to perform the 2D finger trajectories. The second agent acts as the interaction protocol, interpreting and translating to 3D operations the 2D finger trajectories from the first agent. We restrict the learned 2D trajectories to be similar to a training set of collected human gestures by first performing state representation learning, prior to reinforcement learning. This state representation learning is addressed by projecting the gestures into a latent space learned by a variational auto encoder (VAE).
\end{abstract}

\section{Introduction}\label{introduction}
\noindent
The goal in user interface (UI) design is to propose a communication protocol between human users and a given machine that is intuitive, quick, precise, and which minimizes the amount of training required for new users not yet familiar with it. Designing such an interface is not trivial, as some of these desired properties are contradictory. Furthermore, some of these objectives are difficult to quantify, such as intuitivity of the interface or, more generally, user satisfaction.

Our work focuses on a specific component of touch user interfaces, which we call the interaction protocol. This protocol defines the rules that allow the computer to interpret 2D user gestures performed on touch tables into actions in the virtual environment. In the literature, this interaction protocol refers to the software side of an interaction technique \cite{hinckley2012human}\cite{Ortega:2016:IDU:3002573}. In this paper, we address the problem of automatically learning a suitable interaction protocol for graphical user interfaces on touch surfaces, which requires users to manipulate 3D objects, for instance in computer assisted design (CAD) software or in geographic information systems (GIS). In these situations, the problem is particularly ill-posed, as the trajectories produced by a user on the flat touch screen are restricted to a 2D surface, whereas the applications require the user to perform manipulations in a virtual 3D environment. To give a concrete example, inspecting a virtual mechanical product or navigating in a virtual building or city requires the possibility to change the camera viewpoint through rotations, translations, zooming, i.e. to manipulate 6 degrees of freedom (3 for the camera position and 3 for the camera direction) through trajectories of eventually multiple fingers in the 2D plane of the touch table. There is no universally accepted canonical solution for this kind of problem.

%From a naive point of view, we can choose to constrain ourselves to an \emph{active plane} in the 3D environment, defined for instance according to the camera view and a depth parameter, in order to be able to accurately transpose 2D actions into 3D actions. This method is however tedious for the user and is therefore not suitable for a time efficient, satisfying interface.

Advanced methods approach this mapping from gestures to actions in a 3D environment using several parameters: not only the 2D gesture themselves, but also the position of the camera (view of the user), the state of the 3D environment, etc. \cite{bowmanetal06,cashion}. Theoretically, these methods offer more complex manipulation strategies and higher efficiency. However, the challenge here lies in the combination of precision and efficiency on one hand, and ease of use and learnability (by humans) on the other hand.
% according to two criteria: the precision of the action and the ease of use. The criterion of accuracy alone being easy to maximize, it is thus the maximization of the usability while maintaining an acceptable precision which presents the challenge of these interaction techniques; indeed, high level interactions tend to be complicated and are not always easily understood by users.

In this work, we propose to learn these interaction protocols automatically from interactions with humans. The mapping from 2D gestures to actions in the 3D environment is performed by a trainable agent whose policy is learned using reinforcement learning (RL). Such an agent observes user gestures, translates them into actions in the 3D environment and receives a reward, which should be related to user satisfaction in the optimal case. The motivations behind this choice are two-fold:
\begin{itemize}
  \item to automatically learn complex interactions protocols instead of handcrafting them;
  \item to create \textit{adaptive} user interfaces, where not only the human users (classically) adapts to the interface, but the computer also adapts to the way the interaction protocol is imagined by the user through online learning during usage.
\end{itemize}
\noindent
The main challenge lies in the requirement of massive amounts of interactions, necessary for current RL algorithms, but which are difficult or impossible to come by when humans are involved. Requesting users to provide gestures and feedback on satisfaction at each iteration would be overly complicated and not realistic for any complex application.

In this work, we propose to circumvent this problem by firstly pre-training the RL agent from interaction with a learned user model which is jointly learned with the target agent. The interactions of the user model are statistically constrained to natural interactions collected in a static dataset. Secondly, creating loss / reward signals during this pre-training phase from success measures in standard interaction tasks, e.g. ``\textit{go to place X}''.
% firstly building user representations from real data using generative models, and secondly defining learning signals adapted to specific applications.

The paper is organized as follows:
section \ref{ref:related} discusses related work in reinforcement learning and HCI.
Section \ref{sec:monoagent} presents the exact formulation of the HCI problem as a reinforcement learning problem.
Section \ref{sec:multiagent} introduces the main contribution of this work: the formulation as a cooperative multi-agent problem, where the user model is jointly learned with the interaction protocol, restricting simulated interactions to natural ones.
Two different experimental sections report evaluations of the approach on two scenarii of increasing complexity.
Section \ref{ref:toyproblem} describes experiments on a simple 2D environment where users manipulate a 2D object on a 2D surface in a similar fashion to the widely known ``\textit{Pinch-To-Zoom}'' interface developed for smartphones and tablets. The common solution of this type of environment being known, the objective here is to automatically learn this interaction protocol from interactions instead of handcrafting it.
Finally, section \ref{ref:3Dproblem} describes experiments on the targeted application, namely learning a 2D to 3D interface protocol involving navigation in 3D environments. This application features additional complexities, such as the non-unicity of the solution (discussed in Sec.\ref{sec:multiagent}) and the impossibility to analytically define an optimal user.

\section{Related Work}
\label{ref:related}
\noindent
Our work stands between two active fields of computer science: human-computer interface (HCI) and machine learning (ML). We will shortly describe relevant work in both areas to paint the background. \\
\myparagraph{Adaptive user interface} The goal of Adaptive User Interfaces (AUI) is to adapt its visualization and its interactions to fit individual users’ intent better. Machine learning, in this case, is traditionally used for user intent modelization. For an overview of state-of-the-art AUI with adaptive visualization, see \cite{cortes07}. To our knowledge, there is no prior work on learning the interaction protocol of an interface with continous action space.\\
\myparagraph{Reinforcement learning and UI design} Reinforcement learning (RL) is a machine learning framework in which a software agent learns to solve an environment by taking actions that maximizes some cumulative reward. RL has seen some specific uses for AUI design, more precisely for user profiling and representation tasks. In \cite{Seo2000}, an agent learns to detect user preferences implicitly from observing user behavior instead of direct feedback. In \cite{6903138}, an agent uses user feedback to display personalized web pages.\\
\myparagraph{Machine learning and 3D interaction design} 3D user interface (UI) design has been studied for about 20 years \cite{Bowman:2001:IUI:1246684}. In the 3D UI context, a good overview of current state-of-the-art 3D UI methods is given in \cite{Ortega:2016:IDU:3002573}. While ML has been used in user interface and user experience design for about two decades \cite{10.1007/3540634932_3}\cite{Weld:2003}, using machine learning for interaction design is to the best of our knowledge an application yet to be explored. In the UI context, ML is classically used to improve the accuracy of an existing interaction protocol: in \cite{Weir:2012}, a gaussian process regression is used to improve touch accuracy. In \cite{Lu2012}\cite{Debard2018}\cite{Chen2014GM}, supervised deep learning is used to improve the recognition rate of some multi-touch gesture classes.\\
\myparagraph{Reinforcement learning and generative models} Combining the latent representations of generative models with policy learning was explored in some recent work. In \cite{Higgins2017}, the encoding part of a modified VAE is used to build disentangled representations for a RL policy to use in domain adaptation tasks. In \cite{Nair18}, a VAE together with an agent are trained for different purposes: synthesize training data from real observations of the policy, embed the observations to provide latent representations to the policy and measure reward signals in the latent space.\\
\myparagraph{Learning from demonstration} In our paper, we are trying to learn a user model from a small part of all the possible interactions a user can perform. In \cite{DBLP:conf/icml/RossB12}, a model-based agent is built from examples given by a demonstrator that can either be a generative model, open loop excitation or an expert.

\section{Learning an interaction protocol as a reinforcement problem}
\label{sec:monoagent}
\begin{figure}[t] \centering
   \includegraphics[width=0.7\linewidth]{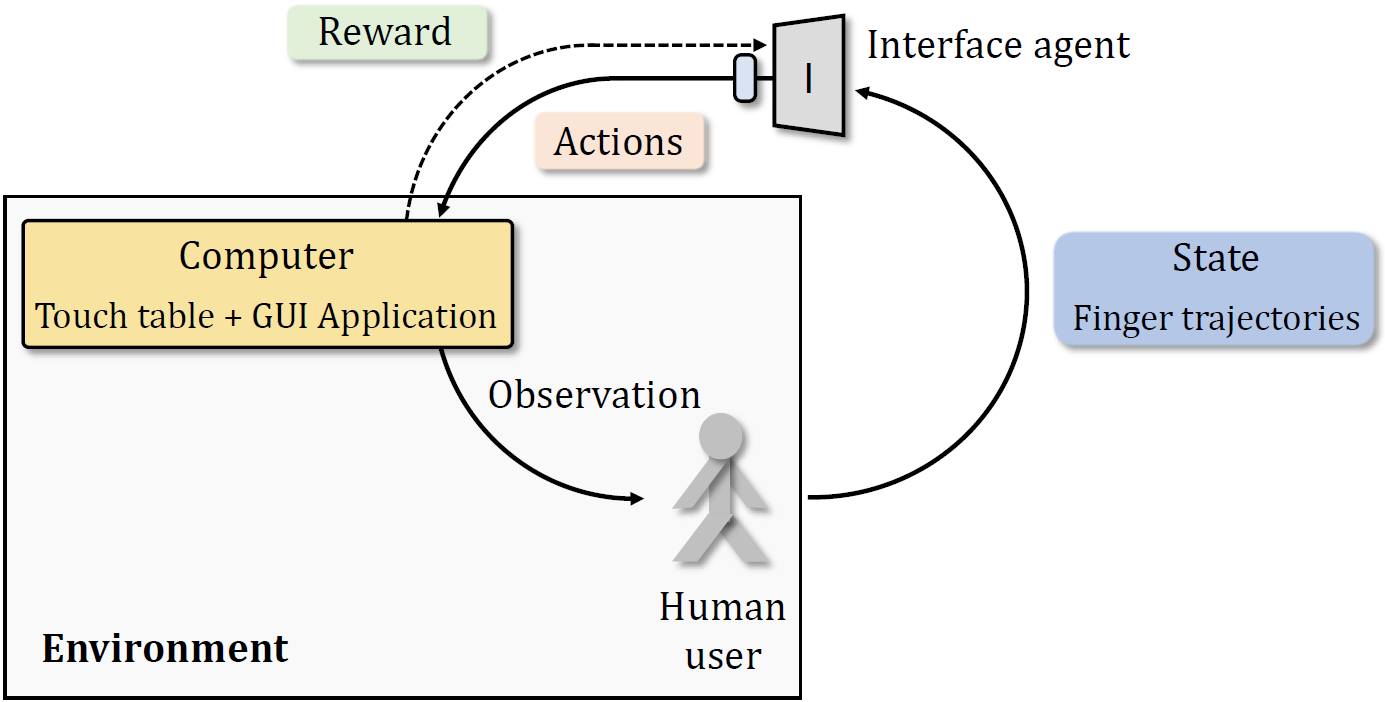}
   \caption{\label{fig:monoagent}Learning user interface protocols from interactions with users as an MDP/RL problem.}
\end{figure}

\noindent
% This section will detail the formalization of our problem into a reinforcement learning one. As stated in the introduction,
Our goal is to learn an interaction protocol coupling user trajectories with actions in a 3D application like CAD or GIS while maximizing the user’s satisfaction. We cast this as an RL problem, where the agent gets observations in the form of finger trajectories and outputs actions, which correspond to viewpoint changes in a 3D environment. Because user intentions and gestures can have long term dependencies and depend on multiple latent factors, this problem could be modeled as a partially observable Markov decision process (POMDP). Assessing the complexity of such a modelization for a novel application, we prefer in this paper to consider the problem as a fully observed Markov decision process (MDP) by stacking two-time instants. This implies considering that the information given at two following time instants is enough to predict user intent. The problem is then treated as an MDP with continuous observation and action spaces. The agent $A$ observes the state in the form of finger trajectories $s$ and receives a reward $r$ after performing an action $a$ in the application. An agent $A$ learns a policy $\pi$ such as $a = \pi(s)$ to maximize its expected return, i.e., the expectation of cumulated reward. Fig. \ref{fig:monoagent} illustrates this situation.

In the RL nomenclature, the agent interacts with an environment, which, in our case, corresponds to, both, the human user and the application, e.g. a CAD or GIS software (see Fig. 1). The agent observes a state, i.e. the user’s finger trajectories, and then performs actions that change the viewpoint in the 3D software and which lead to a new state (new user gestures). The agent also receives feedback in the form of a reward. It is important to note here that user satisfaction is difficult to measure directly if we do not want the resort to solutions which estimate emotions from facial expressions. In the next sections, we will propose proxy metrics which approximate satisfaction.

\section{Jointly learning the interface protocol and human behavior}
\label{sec:multiagent}
\noindent
Deep networks require large-scale training datasets, and Deep RL is not an exception. More so, RL requires dynamic data in the form of interactions, typically millions or billions when observations are of high dimensions and/or when the regularities are complex. In robotics, where interactions with physical robots are slow (not faster than physical time) and expensive, this leads to the tendency of training from simulations, for instance \cite{Beattie2016DeepMindLab}\cite{Edward2019Bench} for robot navigation and \cite{PengAbbeelTransfer2018} for grasping, and to the sim-to-real transfer problems \cite{PengAbbeelTransfer2018}.

Similar to robotics, learning from human interactions is limited. It is restricted to physical real-time, and the effort required from humans during training is to be taken into account. More so, human time is expensive. For these reasons, we address this by simulating the environment, which in our case also involves simulating the human user. However, while simulating robots through handcrafted solutions is feasible, at least approximatively, human behavior is inherently difficult to model. For this reason, we propose a formulation where human behavior is learned jointly with the interface task itself. The next two sub sections describe the two main challenges for this task: (i) restricting the learned user behavior to realistic human gestures (sub section \ref{sec:vae}), and (ii) solving the joint learning problem (sub section \ref{sec:MARL}).

\subsection{Learning the manifold of natural human gestures}
\label{sec:vae}
\noindent
Let $x{\in}\mathcal{X}$ be an observation in the form of natural two-finger trajectories performed by a human user. We define an observation as a $N$-length sequence of 4-tuples, each 4-tuple consists of a pair of coordinates $(x,y)$, one for each of the two fingers. $\mathcal{X}{=}\mathbb{R}^{4N}$ is the space of observations of length $N$. The gesture space $\mathcal{X}$ thus covers all possible pairs of 2D trajectories, including trajectories which are anatomically impossible to perform by human fingers. Our objective is to learn a subspace which corresponds to gestures naturally performed by humans. To this end, we suppose the existence of a training dataset of natural gestures $X=\{x_i\}$, which have been collected from user interactions. This training data can be collected without any manual annotation as simple interaction traces.

% These trajectories do not need to be correlated to the environment the agents will be working on, but a higher variability in the trajectories will give the user model more flexibility toward the definition of a solution.
We want to build a model capable of producing any natural gesture $x$ from a latent representation $z$. Sampling values from $z$ should provide us samples of the manifold of natural human gestures. This involves learning a distribution $p(x|z)$ and to be able to evaluate it from a given $z$. Restricting our simulated user to produce samples of the latent representation $z$ should therefore restrict it to produce natural gestures.

Several approaches exist for learning generative models of probability distributions from training data, among which are Generative Adversarial Networks (GANs) \cite{NIPS2014_5423} and Variational Auto-Encoders (VAEs) \cite{Kingma2014AutoEncodingVB}. Our definition of $p(x|z)$ can be related to the generative part of a GAN or the decoder part of a VAE. In this work, we chose VAEs for two reasons:
they are simpler to train and less sensitive to hyperparameters; and the
 latent space is smoother, due to its soft constraint to be close to a multivariate Gaussian.
 % Their main drawback is the fuzzyness of the generated samples, which is of no concern to us in our case, as sampling is not an issue: the sampling dimension is small (compared to images for example) and small variations in the reconstruction are acceptable to an extent as long as the gesture keeps its semantic.

The VAE is trained on the dataset $X$, approximating the distribution $p_\theta(x)$ by measuring the reconstruction error on a sample $x_i$ coded by an encoder $E$ into a code $z_i$, then reconstructed into $\hat{x_i}$ using a decoder $D$. To describe the problem from a probabilistic point of view, the probability $p_\theta(x)$ of a sample $x$ can be decomposed into a prior and a likelihood as:
\begin{equation}
p_\theta(x) = \int p_\theta(x|z) p_\theta(z) dz
\end{equation}
where the prior on $z$ is defined as a standard Gaussian distribution $
p_\theta(z) = \mathcal{N}(\mathbf{0},\mathbf{I})
$.
In our case (continuous values), we can assume that the likelihood is Gaussian distributed:
\begin{equation}
p_\theta(x|z) = \mathcal{N}(x|D(z,\phi_d),\sigma^2\mathbf{I})
\end{equation}
where $D(z,\phi_d)$ is the decoder of the VAE. The integral is difficult to evaluate, but can be approximated by a point estimate $z{=}q_\phi(x)$ from the variational distribution $q$:
\begin{equation}
p_\theta(x) \approx \mathcal{N}(x|D(q_\phi(x),\phi_d),\sigma^2\mathbf{I})
\end{equation}
$q_\phi(z|x)$ can be seen as an encoder, noted $E(x,\phi_e)$. In this case, we need to ensure that $q_\phi(z|x)$ is a good estimate of the true posterior $p_\theta(z|x)$. This is done using the Kullback-Leibler divergence, noted $D_{KL}$. Considering the approximation error for only a sample $x_i$, the KL divergence becomes $D_{KL}(E(x_i,\phi_e)||p(z))$. As stated earlier, $p_\theta(z) = \mathcal{N}(\mathbf{0},\mathbf{I})$. If we use the $L_2$-norm to measure the reconstruction error, the total error can be written as a variation of the evidence lower bound (ELBO):
\begin{equation}
\label{eq:elbo}
ELBO_i = ||x_i - D(E(x_i,\phi_e),\phi_d)||_2 - \beta \ D_{KL}(E(x_i, \phi_e)||\mathcal{N}(\mathbf{0},\mathbf{I}))
\end{equation}
where $\beta$ is a parameter allowing us to adjust the tradeoff between the reconstruction precision and the latent space regularity \cite{Matthey2017betaVAELB}. We can then update $\phi_e$ and $\phi_d$ by minimizing this error using back-propagation.

\subsection{Cooperative Multi-Agent RL}
\label{sec:MARL}
\begin{figure}[t] \centering
   \includegraphics[width=\linewidth]{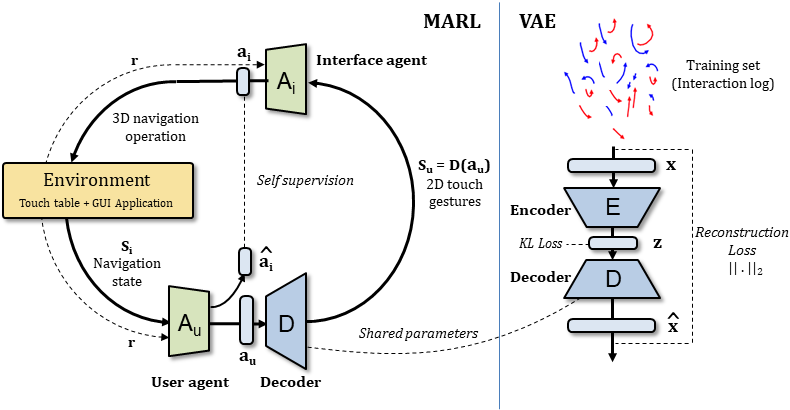}
   \caption{\label{fig:multiagent}Cooperative multi-agent RL problem for jointly learning user interface protocol and user behavior. Generative models are  blue and RL policies are green (best viewed in color).}
\end{figure}

\noindent
We formulate the task of jointly learning the user interface protocol and human behavior as a cooperative multi-agent reinforcement (MARL) problem, as shown in Fig. \ref{fig:multiagent}. Two agents are learned jointly, each with its own policy:
\begin{itemize}
	\item agent $A_i$ corresponds to the user interface. Learning its policy is the original goal of this work, as this agent is responsible for translating 2D finger gestures into actions in the 3D environment (CAD or GIS software).
	\item agent $A_u$ corresponds to the simulated user, with which agent $A_i$ interacts. The only purpose of $A_u$ is to replace human users during the costly pre-training phase. In contrast to $A_i$, $A_u$ is discarded after training.
\end{itemize}
These two agents are trained to maximize the same objective function, sharing the same reward (detailed in section \ref{sec:reward}), which makes this problem a \textit{cooperative} multi-agent problem. Only $A_i$ directly takes action in the virtual 3D environment, whereas $A_u$ acts indirectly by producing the input of $A_i$.

Learning both agents by maximizing the joint reward without additional constraints could naturally lead to degenerate solutions, which are efficient (allow to navigate quickly), but where the gestures exchanged between the two agents are artificial and not easily and naturally doable by humans. For this reason, we restrict the exchange between $A_i$ and $A_u$ to a representation $z$ learned by the VAE described in section \ref{sec:vae}.
%Once the VAE is trained, we need to find a way to use the latent representation of natural gestures built by the VAE in order to provide the interaction protocol agent with coherent training data.
% Such a task can be performed by another reinforcement learning policy.
More precisely, after training the VAE, we discard its encoder. The agent $A_u$ learns a policy on an action space which corresponds to the latent representation $z$. Each  action $z$ is then decoded to a natural gesture $x$ through the decoder of the learned VAE, as illustrated in Fig. \ref{fig:multiagent}. In other words, \textit{the policy of the user agent learns to produce gestures by navigating the latent space of the VAE}.
% which ensures that every observation that the interaction protocol agent receives is taken from an estimation of the natural gestures distribution.
For readability, for the rest of the paper, we will refer to the interaction protocol agent as the interface agent $A_i$, and to the RL agent sampling in the VAE latent space as the user agent $A_u$. The combination of the user agent and the decoder will be called user model $U$.

The method can be more formally described as follows. The task is a sequential cooperative setup where $A_u$ produces the state of $A_i$ and $A_i$ does not get any observation of the virtual environment. In what follows, we denote $s_.^t$ as the state of an RL agent at time $t$, $a_.^t$ as the action at time $t$ and $r_.^t$ as the resulting reward from action $a_.^t$. $\pi_.$ will denote an agent policy. All symbols are indexed by subscripts $u$ or $i$, which stand, respectively, for the agent $A_u$ and $A_i$. Let $\theta^t$ be the state of the software environment at time $t$, for instance the viewpoint in a building, or the 6D pose of a mechanical object in a CAD problem. Then, a given time step $t$ in our sequential cooperative MARL setup will unroll as follows:
\begin{flalign}\label{eq:multiagent}
  s_u^t &= \theta^t \nonumber\\
  a_u^t &= D(\pi_u(s_u^t)) \nonumber\\
  s_i^t &= a_u^t \\
  a_i^t &= \pi_i(s_i^t) = \Delta\theta^t \nonumber\\
  s_u^{t+1} &= \theta^{t+1} \nonumber\\
  r_u^t &= r_i^t = r(a_i^t,s_u^{t+1})\nonumber
\end{flalign}
where $\Delta\theta^t$ is a variation of the parametrization of the object, and $r_u^t{=}r_i^t$ is the joint reward at time $t$. We can note that the constraint on $A_u$ actions defined by $D$ is mandatory for our setup to converge to a cooperative setup. Indeed, supposing that we directly have $s_i^t = \pi_u(s_u^t))$, the policy of one of the agents will degenerate to an identity, effectively lowering the complexity of the problem by getting rid of the intermediate representation between the two agents. One agent will end up observing and taking actions directly in the virtual environment, breaking the paradigm of this setup.

\subsection{Defining the reward function}
\label{sec:reward}
\noindent
% The quality of the learned interface protocol depends on the definition of the reward function.
The goal of this work is to optimize user satisfaction during interactions, which is not easily measurable. We can attempt to empirically break it down to a set of less subjective parts: precision of the interactions, expressiveness, intuitivity and ease of use. The latter two are difficult to measure directly but can be added as learned soft constraints to the agent.
In this work, we make the assumption that the latent representation learned by the VAE from interaction logs encodes intuitivity and ease of use, leveraged by restricting the user agent $A_u$ to an action space defined as the latent representation of the VAE.

%is properly trained, the user model $U$ made of a RL policy using the decoder will naturally express some intuitivity and ease of use in the solution it converges to when learning to maximize performance.
% The last two quantities cannot be directly measured. We either need some prior knowledge to build a model or request real data from human users. Again, we do not have the means to collect enough real data to train these agents. However, we do have a model that (at least partially) implicitly models these notions, the VAE.

The former two (precision and expressiveness) are performance metrics related to the environment the agents are trying to solve. If we restrict ourselves to training from situations where the objective of the HCI experiment is known, these measures can be optimized directly by defining an appropriate reward function. As examples we could imagine asking users questions like ``\textit{Find Waldo in this building by navigating there''} or ``\textit{view the carburator of this V6 engine from above allowing to see inside it}''. The downside to this approach is restricting training to situations with known outcomes and objectives. This still allows learning interfaces in a co-adaptive fashion, in two consecutive stages: a first off-line training stage on a set of ``training users'', followed by an enrollment training phase, where each user is asked to solve custom scenarii to adapt the system to its own interface behavior. It does \textit{not}, however, allow continuous adaptation during usage with unknown objectives.

We will detail our chosen reward functions in the experimental section. It suffices to say at this point, that they measure a distance to the goal in the given user defined task. However, it is important to remember that our true goal is not for the agents to maximize their rewards, which only partially relate to user satisfaction. Convergence of cumulated reward is a necessary condition for a good solution, but not sufficient. Only humans can assess the true quality of these interaction protocols.

\subsection{Stabilizing learning with self-supervision}
\noindent
In the standard formulation as described above, during training, the interface agent $A_i$ learns to interpret actions produced by the user agent $A_u$, while $A_u$ does not get any (unfiltered) information from the interface and as such cannot infer how the interface will interpret a gesture it produces. Furthermore, since the interface policy $\pi_i$ evolves during training, the target for the user agent $A_u$ is unstable, which makes training difficult.

We stabilize training by forcing $A_u$ to approximate decisions taken by the interface $A_i$ in the form of self-supervision. We add a second predictor head to $A_u$, which predicts the output of $A_i$. This predictor shares common layers with the classical predictor of the policy $\pi_u$, which ensures that the learned feature representation benefits both predictors.
% For $A_u$, building a prior of $A_i$ is a major step toward the resolution of the problem. Indeed, the interface is learning along with the user. If the interface changes, a $(s_u, a_u)$ couple that was getting a good reward might now get a bad reward. Updating a prior of the interface along with the standard training can help the user agent to better adapt during training.
% This can be done simply by modifying $\pi_u$ to not only produce $a_u$, but also predict the action to be performed by $A_i$. By using a unique architecture to compute both values, we make sure that the first layers of $\pi_u$ are trained using both signals.
The new predictor is supervised with the real output of the interface agent $A_i$ using the $L_2$ loss
$
\mathcal{L}_e = ||a_i - \hat{a_i}||_2
$.
If $\hat{a_i}$ is the interface action predicted by $A_u$, and $a_i$ is the interface action predicted by $A_i$. In practice, this loss only affects the user actor $\pi_u$, the critic taking no part in this estimation. This is an external training signal added to the RL signal coming from the critic estimation of the state-action value function $Q$. As a note, adding a coefficient to the new loss to control its impact did not yield any meaningful improvements.

\section{Experiments 1: a simple problem --- solving ``Pinch-to-Zoom''}
\label{ref:toyproblem}

\noindent
As a first proof of concept, we will attempt to solve a well-known continuous HCI, for which a handcrafted solution does exist, the goal being to verify whether learning can discover the existing solution.  Our choice here is the well known ``Pinch-to-Zoom'' interface widely used for smartphones and tablets. The name is a misnomer, since the interface not only allows to zoom, but also to translate and rotate the content of surface through 2D gestures made be two fingers.
We suppose that a user performs gestures with exactly two fingers on a touch screen and we investigate the motion between two different time instants. We denote by $\bl = [ l_x \ l_y ]^T$ the screen coordinates of a single finger at the first instant and by $\bl' = [ l'_x \ l'_y ]^T$ the coordinates at the second instant. If we need to explicitly identify a finger, we will index finger $i$ with a superscript as in $\bl^i$ or $\bl'^i$. The coordinates are normalized between $[ 0 \ 0 ]$ (top-left) and $[ 1 \ 1 ]$ (bottom-right).

\myparagraph{The known solution}
\noindent
We will first derive the analytical form of the known solution before describing the experiments learning it. The gestures performed by the user are a combination of translation, rotation and scaling. We suppose that the 2D finger motion on the screen induces the same 2D motion of the manipulated surface, which can be seen as a special case of affine transformation where the shear component is zero. It transforms coordinates $\bl$ into $\bl'$ as
$
\bl' = \bA \bl + \bt
$
where $\bt = [ \bt_x \ \bt_y ]^T$ is the translation component and the rotation+scaling matrix can be calculated from the rotation angle $\alpha$ and the scaling factor $\sigma$ as follows:
\begin{equation}
\bA =
\left [
\begin{array}{rrrr}
\cos \alpha & - \sin \alpha \\
\sin \alpha &   \cos \alpha \\
\end{array}
\right ]
\left [
\begin{array}{rrrr}
\sigma & 0 \\
0 &  \sigma \\
\end{array}
\right ]
=
\left [
\begin{array}{rrrr}
\sigma \cos \alpha & - \sigma \sin \alpha \\
\sigma \sin \alpha &   \sigma \cos \alpha \\
\end{array}
\right ]
\label{ref:A}
\end{equation}
The 4 parameters of the motion are thus $\alpha, \sigma, \bt_x, \bt_y$, which we will combine into a parameter vector $\theta = [ \sigma \cos \alpha \ \ \sigma \sin \alpha \ \  \bt_x \ \ \bt_y]^T$.
If we have motion of two different fingers $(\bl^1, \bl'^1)$ and $(\bl^2, \bl'^2)$, then the following linear relationship between the coordinates and the parameter vector $\theta$ holds:
$
\bd = \bD \theta
$,
where $\bd$ is a vector containing the target coordinates and $\bD$ is a matrix containing the source coordinates in a suitable form:
\begin{equation}\label{eq:affinetrans}
\left [
\begin{array}{c}
l'^1_x \\ l'^1_y \\ l'^2_x \\ l'^2_y \\
\end{array}
\right ]
=
\left [
\begin{array}{rrrr}
l^1_x & -l^1_y & 1 & 0 \\
l^1_y & l^1_x & 0 & 1 \\
l^2_x & -l^2_y & 1 & 0 \\
l^2_y & l^2_x & 0 & 1 \\
\end{array}
\right ]
\left [
\begin{array}{c}
\sigma \cos \alpha \\
\sigma \sin \alpha \\
t_x \\
t_y \\
\end{array}
\right ]
\end{equation}
Because $\bD$ is always invertible (except for the degenerated case where both fingers are at the origin), this linear equation can be solved easily as
$
\hat{\theta} = \bD^{-1}\bd
$.

\begin{figure}[t]
   \begin{center}
   \begin{tabular}{cccc}
   \begin{minipage}{2.8cm}
   \includegraphics[width=3cm]{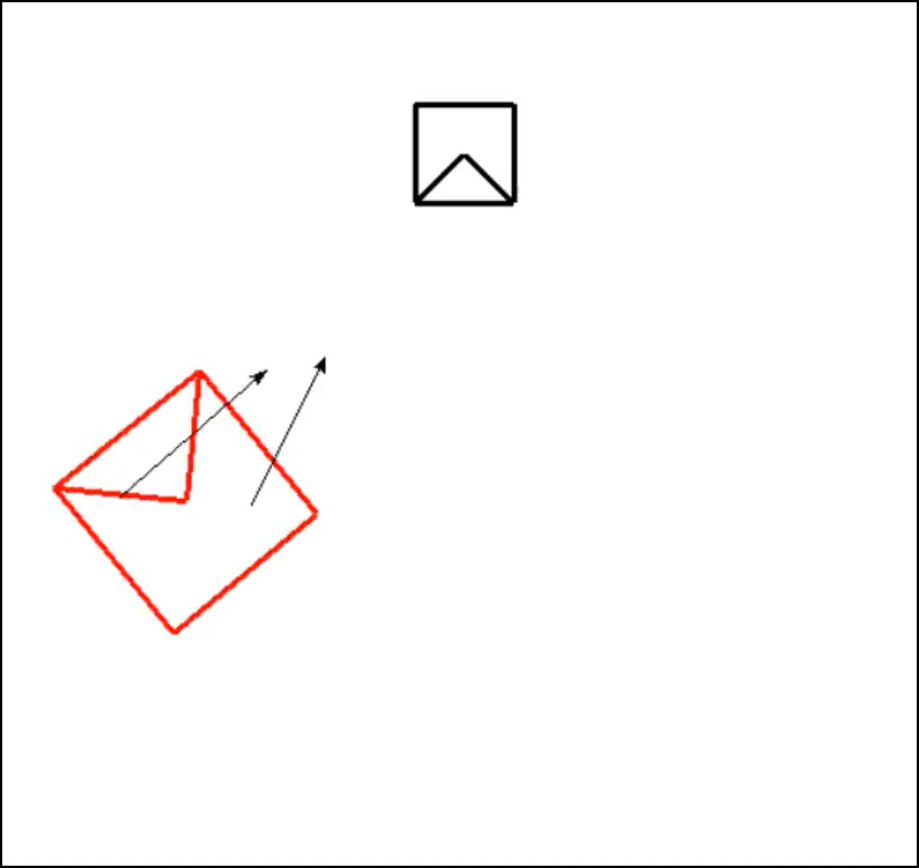}
   \end{minipage} &
   \begin{minipage}{2.8cm}
   \includegraphics[width=3cm]{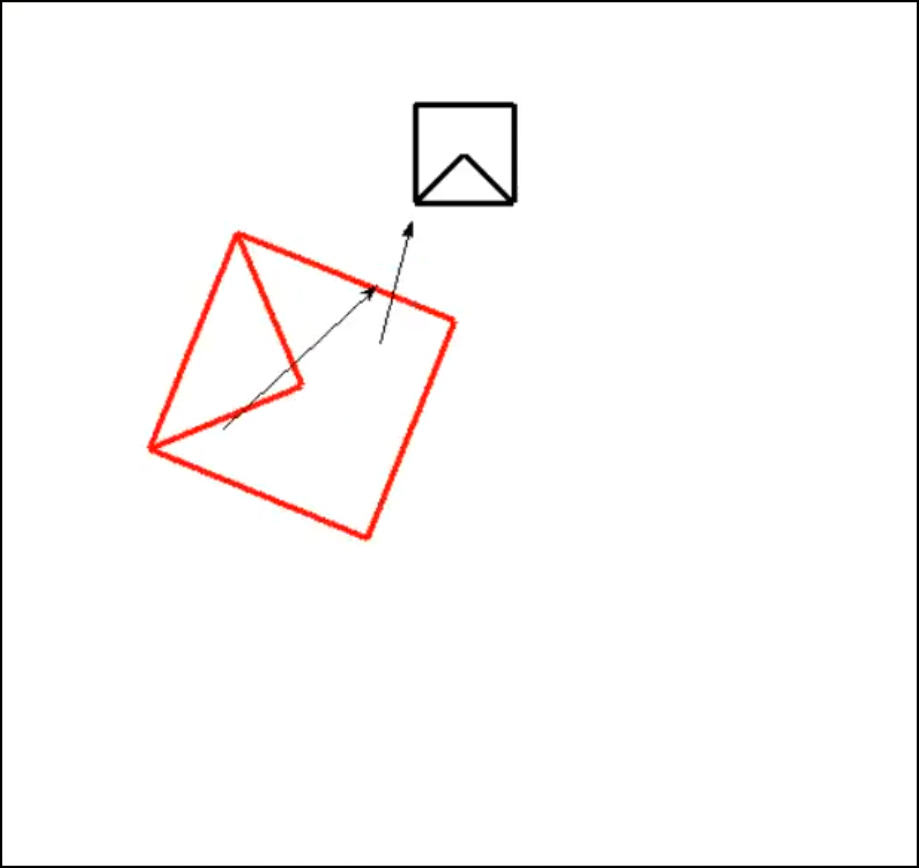}
   \end{minipage} &
   \begin{minipage}{2.8cm}
   \includegraphics[width=3cm]{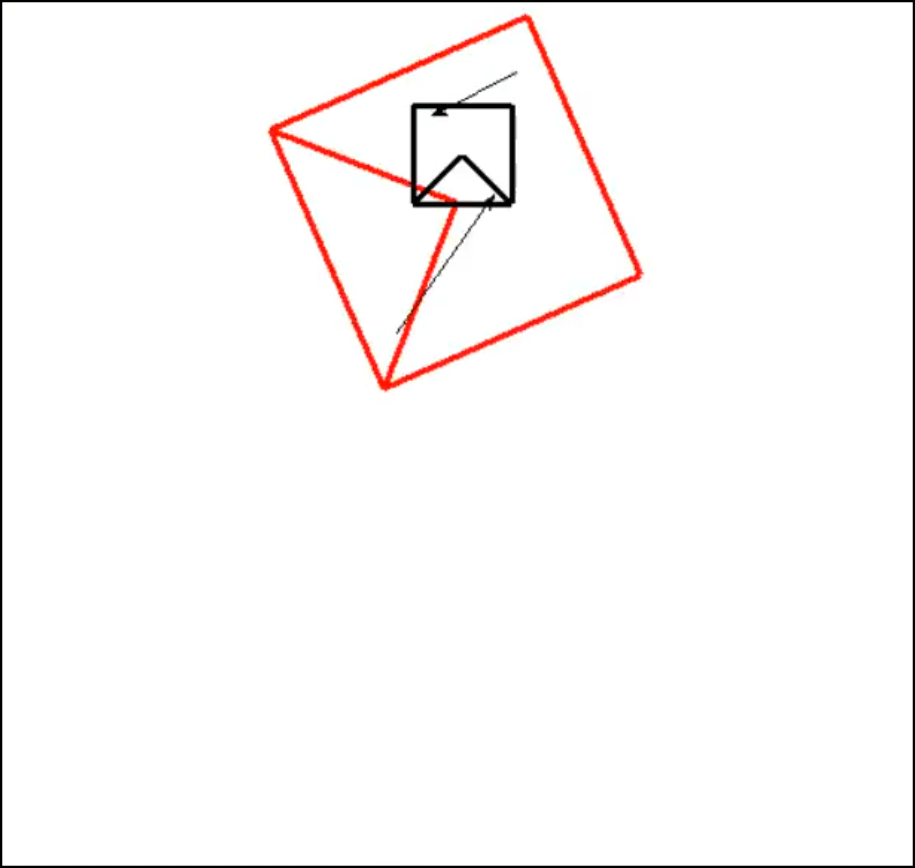}
   \end{minipage} &
   \begin{minipage}{2.8cm}
   \includegraphics[width=3cm]{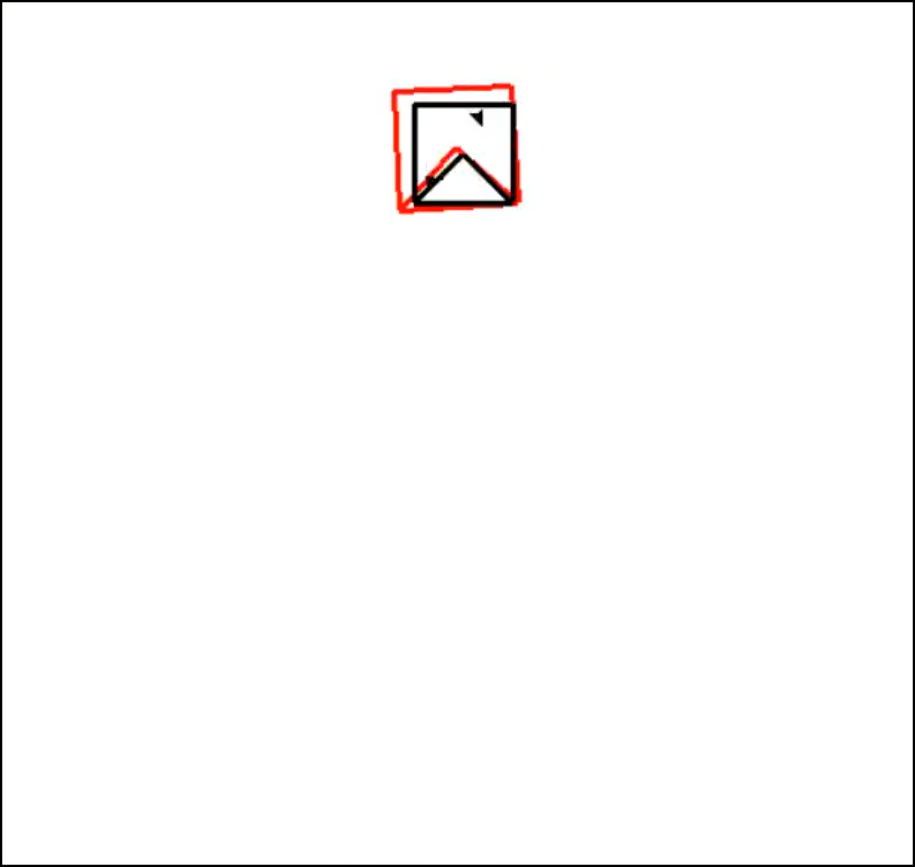}
   \end{minipage}
   \\
   \end{tabular}
   \caption{\label{fig:toyrollout}An example rollout of the interface policy learned for the ``Pinch-To-Zoom'' problem. The rectangle need to be superimposed, finger trajectories are indicated by arrows.}
   \end{center}
\end{figure}

\myparagraph{Learning a solution}
% This can be done in a simple environment where we can define an optimal user action at each time step. The notion of optimal user is important: this ensures that the quality of the reward is directly caused by the quality of the agent actions. In such a case, we do not need a human user nor a complex user model: we can analytically define a user, i.e. one that would maximize the reward if it was using the analytic solution of the interface.
We now let an RL agent learn this protocol. Let the state of the agent be a two-finger motion
$
\bs_i = [ \bl^1_x \ \ \bl^1_y \ \ \bl^2_x \ \ \bl^2_y \ \ \bl'^1_x \ \ \bl'^1_y \ \ \bl'^2_x \ \ \bl'^2_y  ]
$ performed by the user, where superscripts index fingers and subscripts indicate x or y coordinates.
Agent actions
$\ba_i = [ \sigma \cos \alpha \ \ \sigma \sin \alpha \ \  \bt_x \ \ \bt_y]^T
$ are continuous vectors of size 4, which correspond to the parameters of an affine motion transformation without shear component. Note that while this affine transformation has the same functional form as the one expressed in the analytical solution above, we here describe output motion only (motion the manipulated object will endure) and not input finger motion. In other words, in this RL scenario, we do \emph{NOT} suppose that object motion equals finger motion.

The environment is a simple scenario, where a user is required to move a virtual surface containing an object (a red rectangle). The goal is to bring this object to a fixed position by superimposing it on an object which does \textit{not} move with the manipulated surface, i.e. a black rectangle ``painted'' on the glass of the device. The reward function in this task is the sum of $L_1$- and $L_2$-distances the object vertices and the target vertices:
\begin{equation}
  r = -\sum_i{(\ ||\bo_i - \bt_i||_1 + ||\bo_i - \bt_i||_2\ )} -0.2
 \label{eq:reward}
\end{equation}
where $\bo_i$ and $\bt_i$ are the coordinates of $i-th$ vertice of the respective rectangle. Let us recall that the interface agent does \emph{not} have access to these positions, else it would learn to simply ignore user gestures. The constant $-0.2$ reward is set to continuously encourage fast solutions. A positive reward of $+25$ is given if the agent successfully finishes an episode. These arbitrary values are chosen so that the expectation of the sum of rewards per episode is close to 0 for an agent close to the optimal solution.
% Environment knowledge for the interface is valid only if the problem needs to be modelled as a Partially Observable MDP (POMDP) and observation alone of the environment is not enough to decide which action to perform.

% \begin{figure}[t]
%    \begin{center}
%    \includegraphics[width=15em]{./figs/2Denvironment.png}
%    \caption{\label{fig:2Denv} 2D environment. The movable object is the red shape, the target position is the black one. Arrows display user action at the current timestep.}
%    \end{center}
% \end{figure}

\myparagraph{Handcrafted simulation of the user}
For this toy problem, the cooperative multi-agent formulation proposed in section \ref{sec:multiagent} is not necessary. We instead handcraft a solution simulating a user who is aware of the ``Pinch-To-Zoom'' protocol, i.e. of the known analytical solution. We would like to stress that the agent is of course \textit{not} aware of the solution. The interface solution expressed in \ref{eq:affinetrans} allows us to compute simulated user trajectories: this can simply be done by considering two diagonally opposed object vertices $v_1$ and $v_2$ and their target position $v'_1$ and $v'_2$. For a sampling time, we can consider that the user will move the object toward the target while keeping the vertices $v_i$ on the segments $[v_i,\ v'_i]$ (which is the optimal way to solve the task). It means we can find intermediate positions of $v_i$ on these segments using the linear combination:
\begin{equation}
v^{inter}_i = (1-\mu) v_i + \mu v'_i \ , \ \mu = \max(1, \frac{0.5}{||v'_i - v_i||_2})
\end{equation}
$\mu$ is the user's gesture velocity. A small $\mu$ will mean small relative increments toward the target. This definition of $\mu$ goes in the sense that a human user will tend to do faster gestures while far from the target and slower, more precise gestures while close to it. Now that we have two points of the wanted intermediate object position, we can solve the equation \ref{eq:affinetrans} in order to get the transformation of every point of the object to the intermediate position. At last, we can choose two random points $p_1$ and $p_2$ on the object, transform them using the computed $\hat{\theta}$ parameters and build the two trajectories $[p_1, \ p^{inter}_1]$ and $[p_2, \ p^{inter}_2]$. The state $bs_i$ of the agent will be the concatenation of these two trajectories, resulting in a vector of size 8.\\
\myparagraph{Results}
We compare our trained agent to an optimal solution. This optimal solution is easily modelled as we defined both the analytic solution of a user and of the interface. On an average of 100 episodes, the optimal solution finishes an episode in 40 steps and obtains a reward of +0.5 per episode.

After a training of about 300k steps, the interface agent obtains very similar results: on an average of 100 episodes, it finishes an episode in 41 steps and obtains a reward of +0.4. It is visually impossible to separate the optimal solution from the learned agent. An illustration of a rollout in the environment is given in Fig. \ref{fig:toyrollout}.

% \myparagraph{Solving Pinch-to-Zoom with cooperative RL}
% \noindent
% \cw{Rédiger à la fin, pas sur qu'on le garde dans l'article.}
% For this part, we will consider the same problem, environment and notations as before. The major difference is now that we use our model $U$ instead of the handcrafted gesture simulator to produce user gestures. We want to test that $A_u$ can successfully learn to produce natural finger trajectories while $A_i$ and $A_u$ are training all together. $A_i$ must of course find again the same interaction protocol solution.

% \paragraph{Experimental results}
% Description de l'architecture, des hyper-paramètres, du protocole et des métriques. Faire un tableau en fonction du bruit utilisé. Comparaison avec un setup optimal (entièrement handcrafté), courbes tensorboard.
% \qd{mettre la floppée d'hyperparamètres en supplementary? Il y en a vraiment beaucoup.}
% \begin{table}[t] \centering
%     \begin{center}
%         \begin{tabular}{cccccc}
%             \arrayrulecolor{cwblue1} \toprule
%             & & Nb. successful ep. & Mean reward/episode & Mean nb. steps/ep. & Nb. training steps\\
%             \arrayrulecolor{cwblue1} \toprule
%             A & Mono agent no noise & - & - & - & -\\
%             B & Mono agent param. noise & - & - & - & -\\
%             C & Mono agent OU noise & - & - & - & -\\
%             \arrayrulecolor{cwblue1} \toprule
%             D & MARL & - & - & - & -\\
%             \arrayrulecolor{cwblue1} \bottomrule
%         \end{tabular}
%     \end{center}
%     \caption{Toy example results}
%         \label{2Dresults}
% \end{table}

\begin{figure}[t] \centering
   \includegraphics[width=\linewidth]{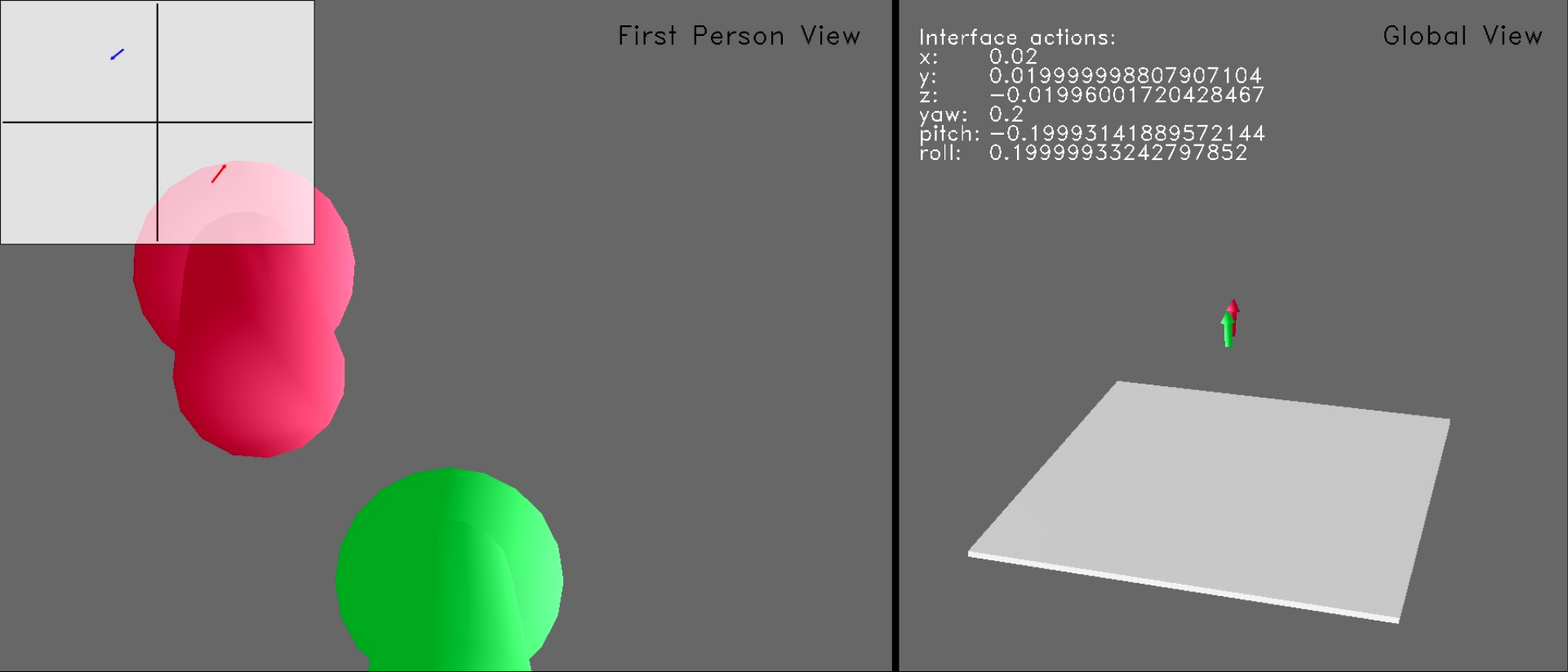}
   \caption{\label{fig:3DEnv}The 3D navigation user interface. Left: the user/camera view; Right: static bird's eye view. Current user gestures are displayed on top left. The goal is to superimpose the green arrow attached to the camera with the red non moving target arrow.}
\end{figure}

\section{Experiments 2: 3D navigation from 2D gestures}
\label{ref:3Dproblem}

\noindent
We now discuss the experiments on the real 3D navigation user interface, for which no optimal solution is known to exist. As described in the introduction, we want to learn an agent to map 2D finger gestures to motion in a 3D environment, an ill-posed problem. To this end, we extended the 2D toy problem to 3D, maintaining the user's goal of moving a (now 3D) content to superimpose an object over a non-moving object, as shown in Fig. \ref{fig:3DEnv}. As there is no simple handcrafted way to simulate a human user, we use the multi-agent RL setup described in Sec. \ref{sec:multiagent}.
% Again, the interface agent needs to assume that the user is performing optimal finger trajectories. This statement is valid in the sense that the user agent will converge toward an optimal solution with respect to the reward function.
The 3D affine transformation to be learned by the interface agent can be expressed  using homogeneous coordinates in 4 dimensions as a $4{\times}4$ matrix $\phi$:
\begin{gather}
  \phi=
  \begin{bmatrix}
    r_{11} & r_{12} & r_{13} & t_1\\
    r_{21} & r_{22} & r_{23} & t_2\\
    r_{31} & r_{32} & r_{33} & t_3\\
    0 & 0 & 0 & 1
  \end{bmatrix}
  , R =
  \begin{bmatrix}
    r_{11} & r_{12} & r_{13}\\
    r_{21} & r_{22} & r_{23}\\
    r_{31} & r_{32} & r_{33}\\
  \end{bmatrix}
  , t =
  \begin{bmatrix}
    t_1\\
    t_2\\
    t_3\\
  \end{bmatrix}
\end{gather}
As we do not consider scaling (redundant with forward motion), the matrix is totally parametrized by 6 coefficients: $[\tau_x,\tau_y,\tau_z,\rho_x,\rho_y,\rho_z ]$, where $\tau_.$ and $\rho_.$ are, respectively, translation coefficients and Euler angles on the 3 axis. We limit the Euler angles to $ \left] -\pi, \pi \right] $ to ensure their unicity given an axis. Such a vector can define transformations as well as object positions when using a fixed referential in the environment.
With this formalization, The user agent $A_u$ gets as observations the camera position vector and learns a policy over actions which are trajectory vectors $[ l^1_x \ l^1_y \ l'^1_x \ l'^1_y \ l^2_x \ l^2_y \ l'^2_x \ l'^2_y]$. The interface agent $A_i$ observes the output of $A_u$ and learns a policy over residual transformation vectors of the camera, i.e.  $s_u^{t+1} = s_u^t + a_i$. We define the reward equivalent to the 2D problem in section \ref{ref:toyproblem}, given in equation (\ref{eq:reward}), the difference being that each object has 2 vertices instead of 3, and that vertices are in 3D space.

\begin{figure}[t] \centering
   \includegraphics[width=1\linewidth]{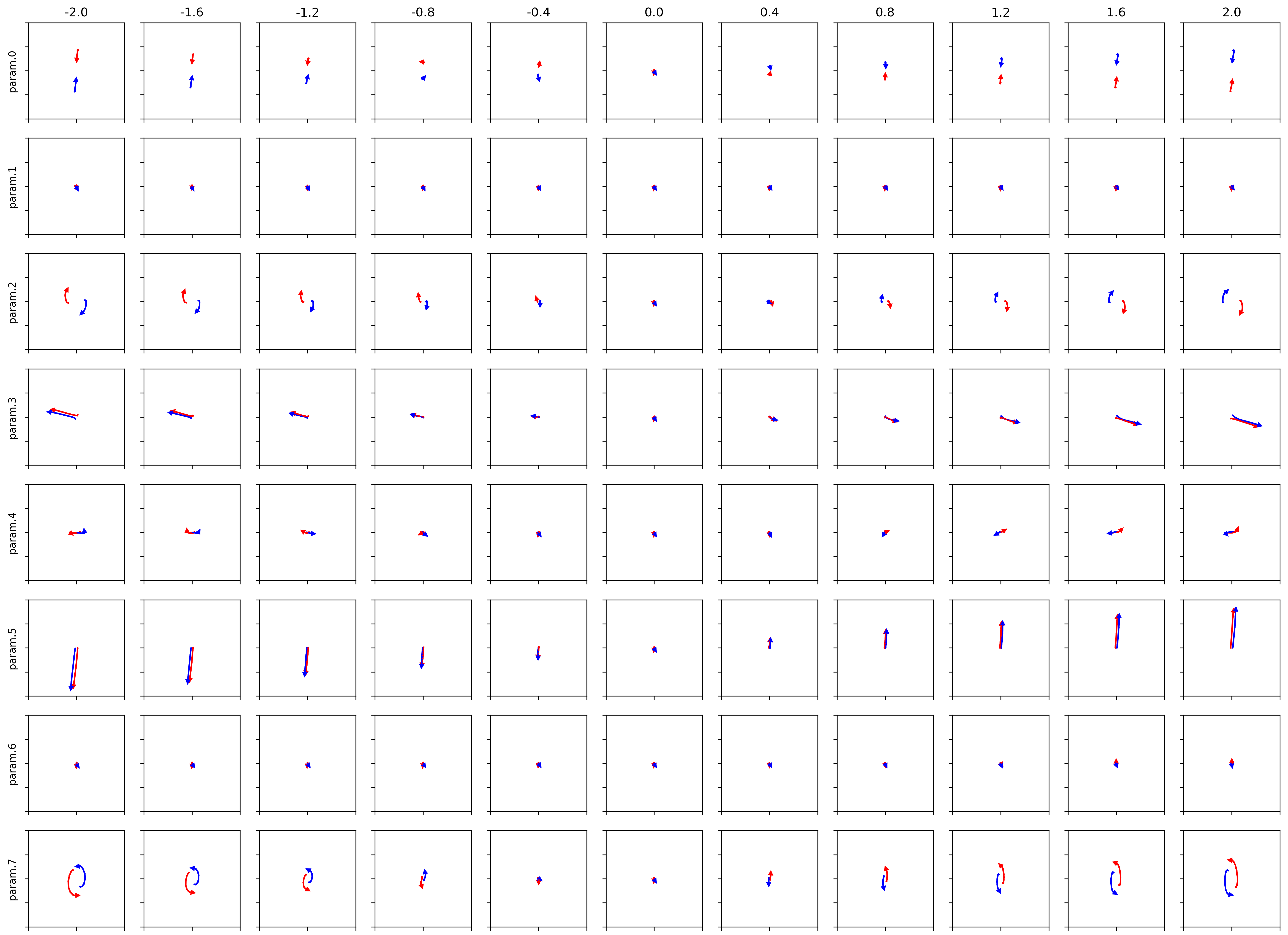}
   \caption{\label{fig:latentspace}Navigating the latent space learned by the VAE. Middle column: zero code $z$. Lines correspond to different modified latent variables (dim=8). Columns correspond to different values.}
\end{figure}

\begin{figure}[t] \centering
   \includegraphics[width=1\linewidth]{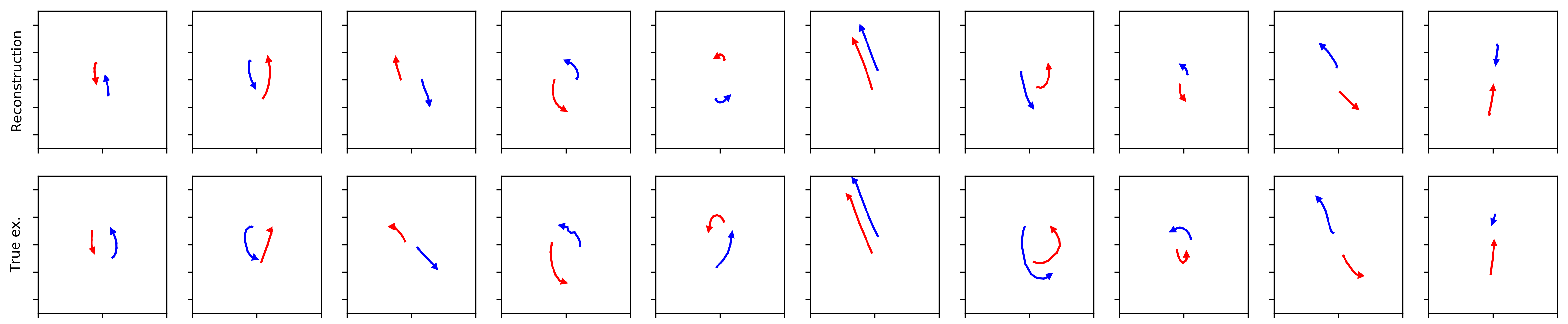}
   \caption{\label{fig:reconstruction}Reconstruction examples made by the VAE. The first line displays reconstructed gestures while the second displays the corresponding original gestures.}
\end{figure}

\myparagraph{VAE training and architectures}
We train the latent 2D gesture representation with a VAE on a dataset of multi-touch interaction gestures, in particular the Itekube-7 Dataset \cite{Debard2018}. We used all the two-finger gestures from the translation, pinch and rotation classes. Each gesture of the dataset was sampled using the dynamic sampling described in \cite{Debard2018} to fix the length of the gestures to 10 timesteps.
The VAE was trained for 50 epochs with $\beta{=}0.07$, a batch size of 128, and a learning rate 0.002.
The latent code is of size 8. The encoder and the decoder are both recurrent. Encoder: a one-layer GRU \cite{cho-al-emnlp14} with a hidden state of size 256 and ReLU activation. It is recurrent in time and reads inputs of size 4 (two 2D finger positions). Second, two FC layer predict, respectively, the mean and the stddev of the latent code from the last hidden state.
The decoder is a two-layer GRU: The first layer has a hidden state of size 128 with  ReLU activation, the second layer has a hidden state of size 4 (to reconstruct the position of both fingers at each timestep), no activation. The entire code is fed to it at every timestep, i.e. the number of unrollings of the GRU will determine the number of timesteps of the reconstruction. At training time, this number is set to the number of timesteps of the original data (10), but at inference time, this number can be arbitrarily, which allows to produce shorter or longer trajectories from the same latent code.

Fig. \ref{fig:latentspace} visualizes the latent representation, and reconstruction examples are given in Fig. \ref{fig:reconstruction}. The representations from the latent space are satisfying as we can observe some high level features and disentanglement: for instance, dimension 0 expresses pinch, while dimension 2 expresses clockwise rotation.
% In the reconstructions, it is important to observe that the semantic of the reconstructed gesture is very similar to the original one.

\myparagraph{Multi-agent RL training}
We chose the model-free off-policy actor-critic method Deep Deterministic Policy Gradient (DDPG) \cite{DBLP:journals/corr/LillicrapHPHETS15}.
In our setup, an epoch cycle consists of two phases: (i) a rollout phase on an episode, where all quadruplets $[$state, action, reward, new state$]$  are stored in the replay memory of the agents; (ii) a training phase where quadruplets are randomly sampled from the memory, batched (in sizes of 4096) and used to train the actor and the critic of the agents. We define an epoch as 100 epoch cycles. The training is arbitrarily stopped when no improvement on the metrics is observed. A training session takes about 2 days on a Titan-X Pascal GPU.

Join training of both, $A_u$ and $A_i$, was not successfull. We suspect the added variance and the moving value of state-action pairs for both agents as a source of the problem. Similar to \cite{Boutilier:TARK96}, we chose to train them in an alternating manner: during an epoch, only one agent will be trained, while the weights of the other agents are kept fixed.

\myparagraph{Stacking timesteps} We consider two ways for the two agents to communicate. The simplest one is a two-instant communication: the decoder produces a gesture of only two time instants, and the interface produces the corresponding action. It is simple, but leads to non-smooth user gestures difficult to appreciate from a human viewpoint is hard. We also consider stacking time instants: $A_u$ produces a complete gesture of 10 timesteps, and $A_i$ must produce the corresponding sequence of actions (9 if there are 10 timesteps). In this case, a step from the RL perspective will contain 10 update steps of the environment. This decouples the update speed of the agents from the sampling speed of the finger gestures, referred to as ``\textit{Stacking}'' in Table \ref{3Dresults}.

\begin{table}[t] \centering
    \begin{center}
        \begin{tabular}{lccc}
            \arrayrulecolor{cwblue1} \toprule
              \ & Mean reward/ep. & Mean \#steps/ep. & Nb. training steps\\
            \arrayrulecolor{cwblue1} \midrule
            No Stacking &  3.6$\pm$1.0 & 53$\pm$1 & 17.6M$\pm$0.4M\\
            + Stacking & 0.5$\pm$1.5 & 56$\pm$4 & 24.6M$\pm$6.3M\\
            \arrayrulecolor{cwblue1} \midrule
            Theoretical Opt.  & 5.0 & 40 & N/A\\
            \arrayrulecolor{cwblue1} \bottomrule
        \end{tabular}
        % \begin{tabular}{lccccc}
        %     \arrayrulecolor{cwblue1} \toprule
        %     & Stacking \ & Self-supervision \ & Mean rew./ep. & Mean \#steps/ep. & Nb. training steps\\
        %     \arrayrulecolor{cwblue1} \midrule
        %     Baseline & - & - & 3.6$\pm$1.0 & 53$\pm$1 & 17.6M$\pm$0.4M\\
        %     +Self Supervision & - & \myyes & 1.3$\pm$0.2 & 50$\pm$3 & 19.2M$\pm$1.1M\\
        %     \arrayrulecolor{cwblue1} \midrule
        %     +Stacking & \myyes & - & 0.5$\pm$1.5 & 56$\pm$4 & 24.6M$\pm$6.3M\\
        %     Ours & \myyes & \myyes & 0.2$\pm$1.2 & 54$\pm$1 & 33.1M$\pm$2.1M\\
        %     \arrayrulecolor{cwblue1} \midrule
        %     Theoretical Opt. & N/A & N/A & 5.0 & 40 & N/A\\
        %     \arrayrulecolor{cwblue1} \bottomrule
        % \end{tabular}
    \end{center}
    \caption{\label{3Dresults}Results on the 3D environment. The last line gives theoretical optimal results based on the interface action amplitude, without considering a naturalness constraint. The Mean reward/ep. is the mean cumulated reward per episode obtained in the best performing epoch. The Mean nb. steps/ep. is the mean number of timesteps needed to successfully finish an episode in the best performing epoch. The Nb. training steps is the number of environment steps that was needed to attain the best performing epoch. Stacking improves usability but \emph{NOT} efficiency.}
\end{table}

\myparagraph{Architectures} In what follows, FCX refers to an FC layer with X hidden units, with layer normalization and  ReLU activation. The actor of $A_u$ is an MLP with two hidden FC100 layers. The output layer is FC and activated with tanh, predicting a vector of size 8 (the latent code $z$ expanded by the VAE decoder $D$). Another FC100 layer is plugged to the first hidden layer, with an output layer producing the estimate $\hat{a_i}$. The critic of $A_u$ is an MLP with two hidden FC100 layers. The policy action is concatenated to the first hidden layer. A linear FC layer predicts the value $Q$.

The actor of $A_i$ is an MLP with two hidden FC64, and an output layer with tanh activation. The output size is either 6 for the standard solution or 6x9=54 for the stacked solution. The critic of $A_i$ has the same architecture as the critic of $A_u$, except hidden layers are FC64.

\myparagraph{Results}
Quantitative results are given in Table \ref{3Dresults}. Each setup was reproduced with 3 different random seeds. We consider that a run has converged whenever all 100 episodes of an epoch are successful. Once it has converged, it can still improve by solving episodes faster. This is measured as the mean number of steps needed to solve an episode. The mean reward per episode is also correlated to the quality of the interaction protocol, but should be interpreted differently. Indeed, a run with a lower mean step per episode but a higher mean reward per episode is most likely less satisfactory than a run with higher steps but lower reward. This is because the first type of solutions tend to be less continuous with harsher action changes, while the second is technically slower but goes in the direction of the objective more smoothly.

\textit{``Stacking''} and usability --- the interaction protocols must also be observed visually in order to assess their global quality: good interfaces should display distinctive characteristics, such as a similar curvature between 2d trajectories and 3D movements of the camera, or well defined classes for similar actions. While stacking does \textit{NOT} improve efficiency (as shown in table \ref{3Dresults}), it makes the protocol usable.
The continuous aspect of gestures using instant stacking is illustrated in \ref{fig:rollout3D}, videos are provided in the supplementary material.
\begin{figure}[t]
   \begin{center}
   \begin{tabular}{cccc}
   \begin{minipage}{2.85cm}
   \includegraphics[width=3.1cm]{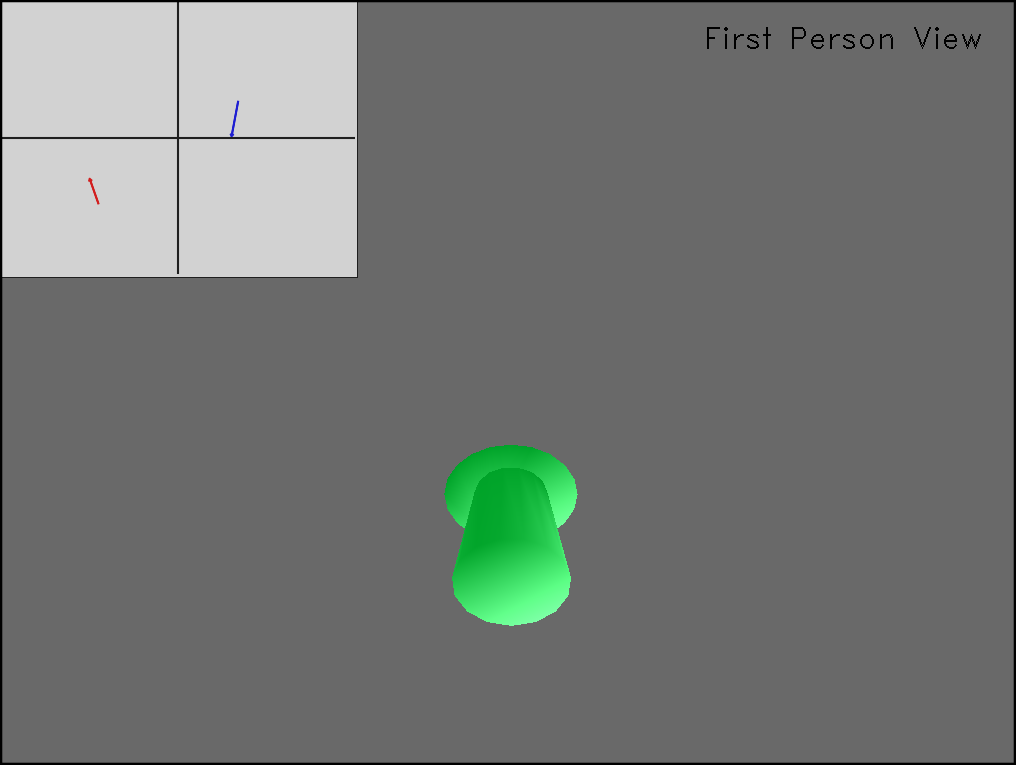}
   \end{minipage} &
   \begin{minipage}{2.9cm}
   \includegraphics[width=3.1cm]{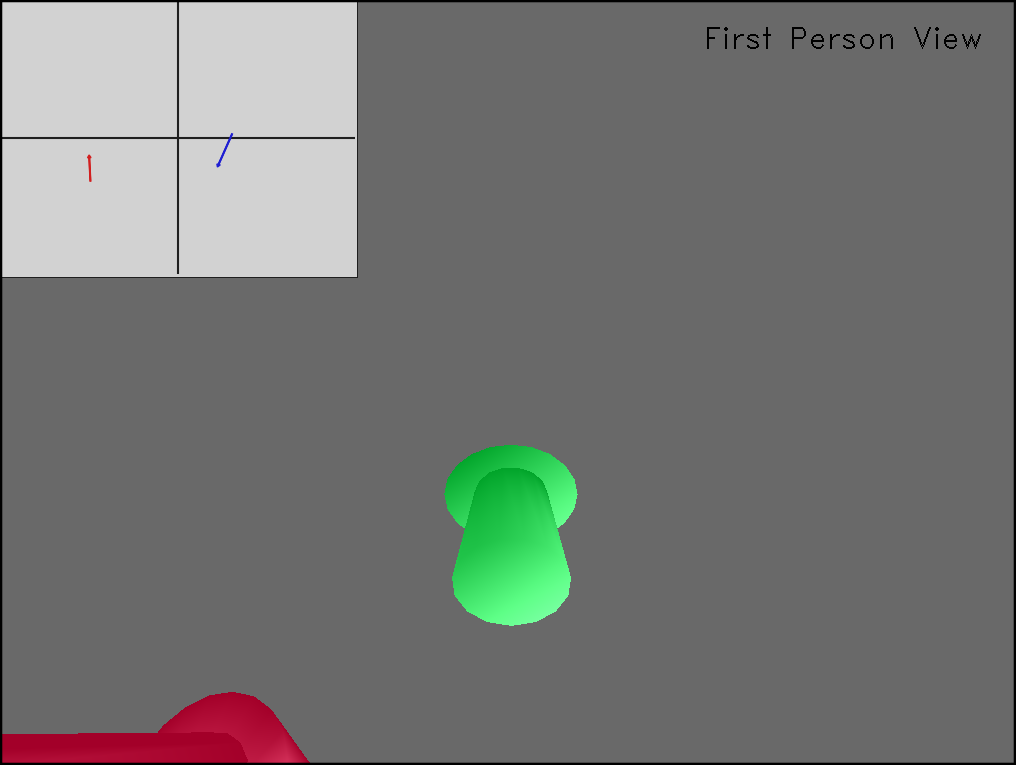}
   \end{minipage} &
   \begin{minipage}{2.9cm}
   \includegraphics[width=3.1cm]{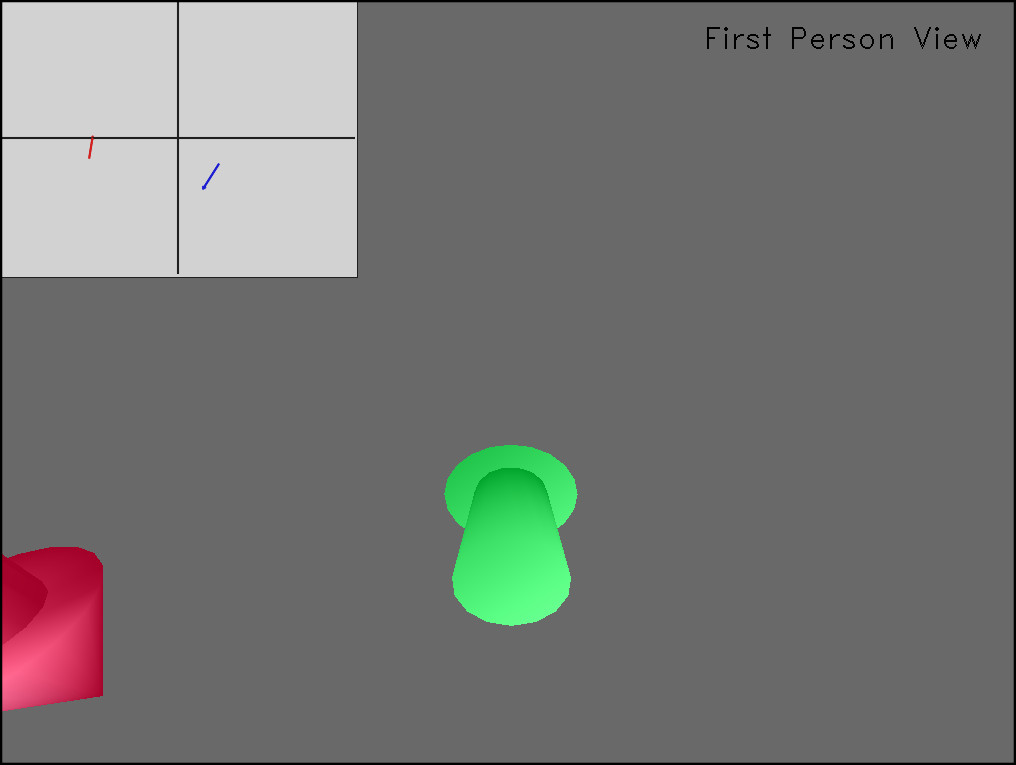}
   \end{minipage} &
   \begin{minipage}{2.9cm}
   \includegraphics[width=3.1cm]{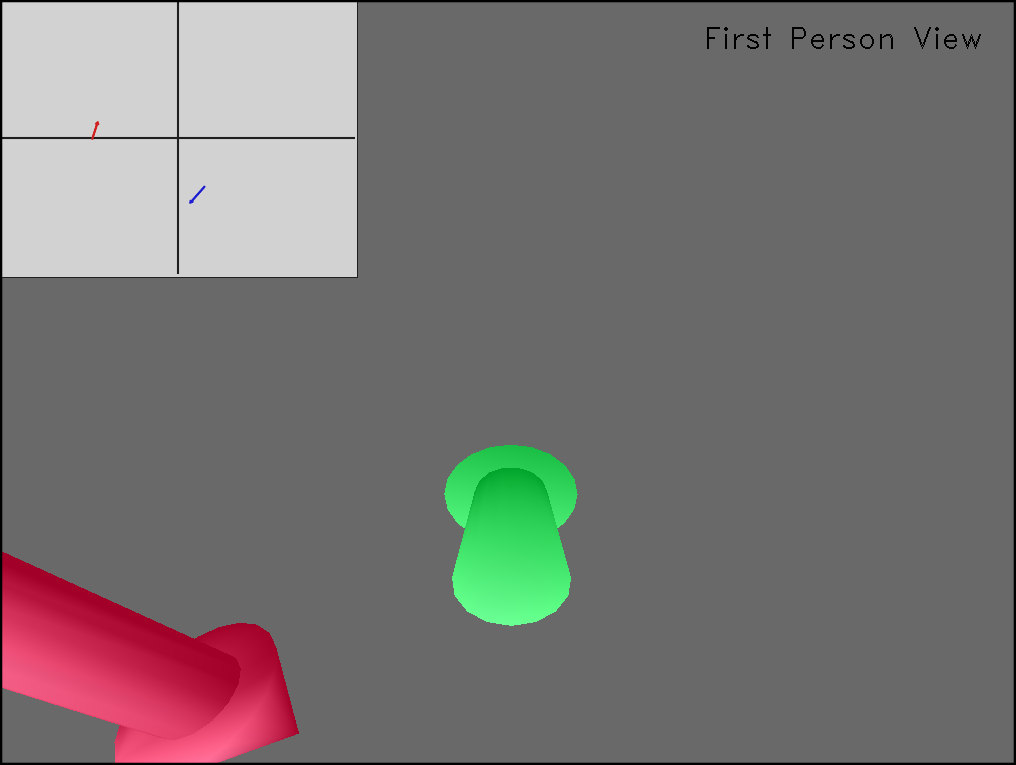}
   \end{minipage}
   \end{tabular}
   \caption{\label{fig:rollout3D}An example of 4 back to back frames from the MARL setup learning the 3D navigation problem with instant stacking. We want to emphasize on the continuous aspect of user gestures (top-left) and the semantic. We can see that the user agent is currently performing a rotation-like gesture.}
   \end{center}
\end{figure}
% \myparagraph{Reward function}
% The most straightforward reward function is a sparse one: $+1$ if the episode is successful, $-1$ if it fails. Because we want the agents to succeed while minimizing the amount of time needed, we will change the negative failure reward as a continuous negative reward, say $-0.1$ at each time step. This will encourage the agents to finish episodes more quickly. While such a reward is theoretically enough, in practice the rarity of valuable learning signal (when an episode succeeds) at the beginning of the training makes the convergence of the agents statistically unlikely. Adding a continuous reward depending on the environment state is a common way to overcome this scarcity. The continuous reward functions used in this paper are described in the corresponding sections of each environment.

\section{Conclusion}
\label{sec:conclusion}
\noindent
We presented a novel method for automatically learning interaction protocols from natural interactions. While our application for this paper is limited to touch interfaces, this setup can virtually be applied to any technology and any software, as long as a large enough interaction dataset can be collected. We want this work to be a step toward a better co-adaptive relation between the human and a computer, allowing for individually suited interfaces and higher levels of interaction.
Future work will model the interaction protocol as a POMDP, which should allow to better represent long term dependencies and tackle more complex regularities.
The main remaining problem is to fine-tune the learned models on real human users, which requires large-scale efforts with a large amount of human partners.

%\begin{acknowledgements}
%\end{acknowledgements}

\bibliographystyle{splncs03}
\bibliography{references.bib}

\newpage
\appendix
\section{Appendix}
\begin{figure}
   \begin{center}
   \begin{tabular}{cccc}
   \begin{minipage}{2.85cm}
   \includegraphics[width=3cm]{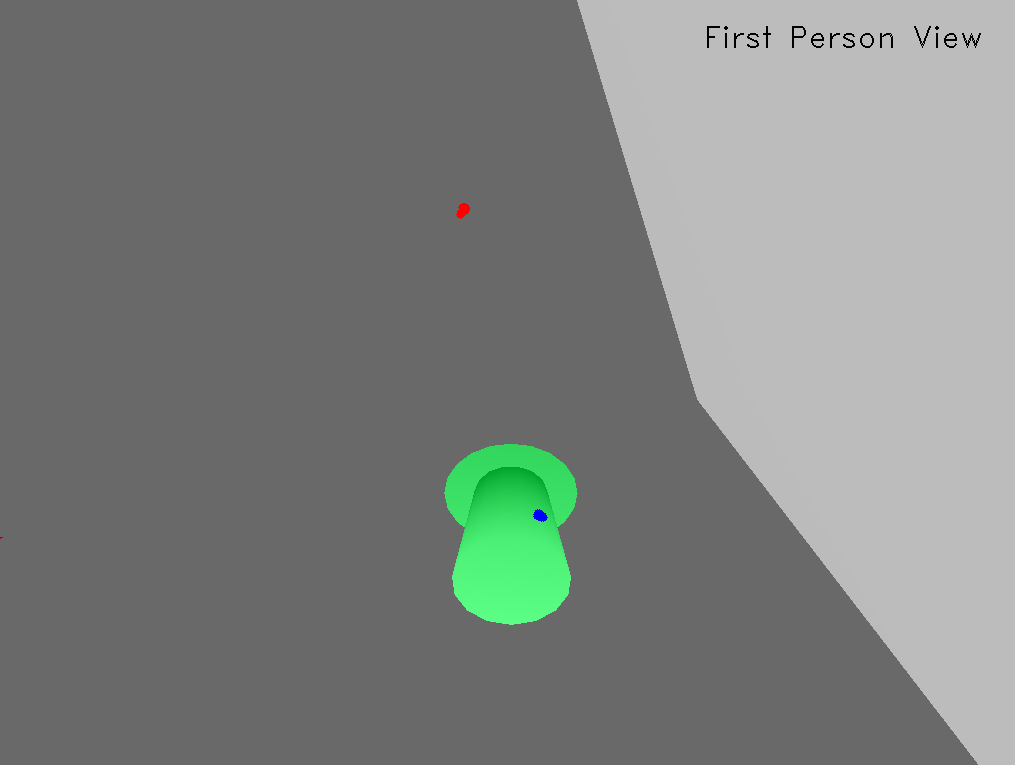}
   \end{minipage} &
   \begin{minipage}{2.9cm}
   \includegraphics[width=3cm]{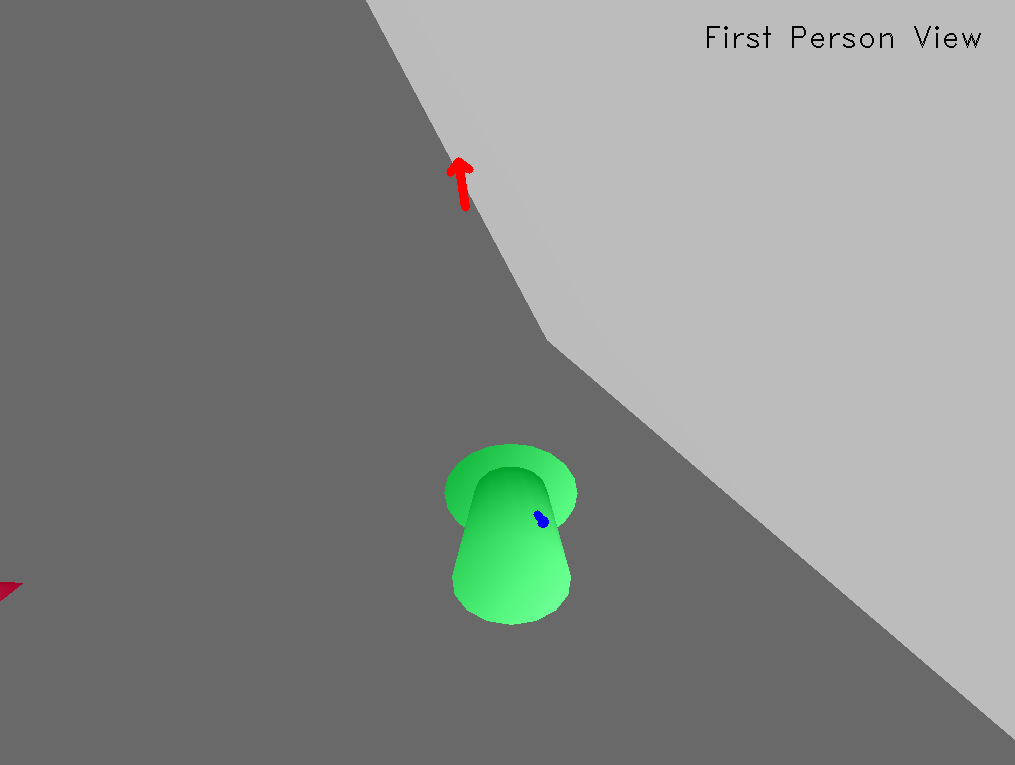}
   \end{minipage} &
   \begin{minipage}{2.9cm}
   \includegraphics[width=3cm]{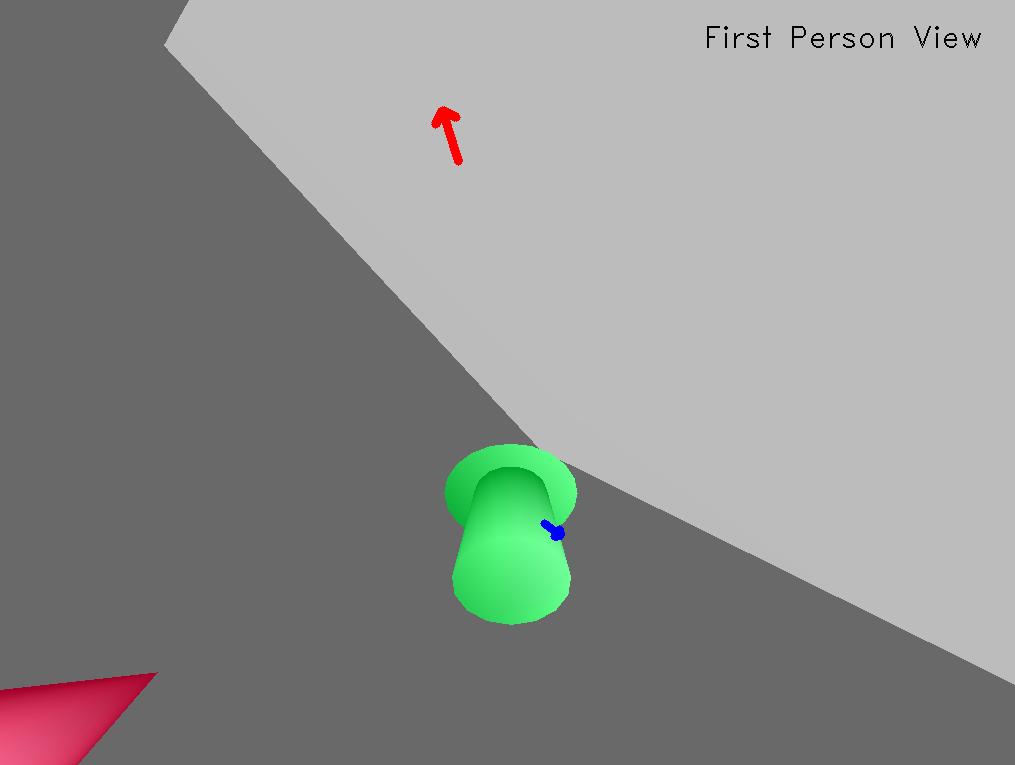}
   \end{minipage} &
   \begin{minipage}{2.9cm}
   \includegraphics[width=3cm]{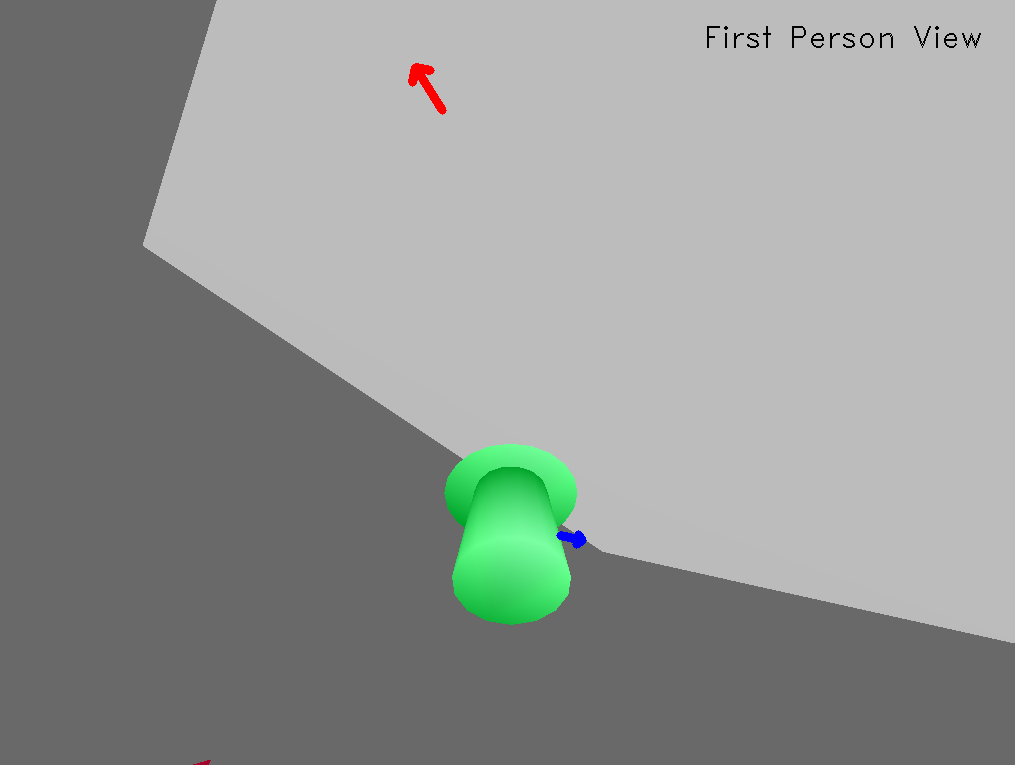}
   \end{minipage}\\
   \begin{minipage}{2.85cm}
   \includegraphics[width=3cm]{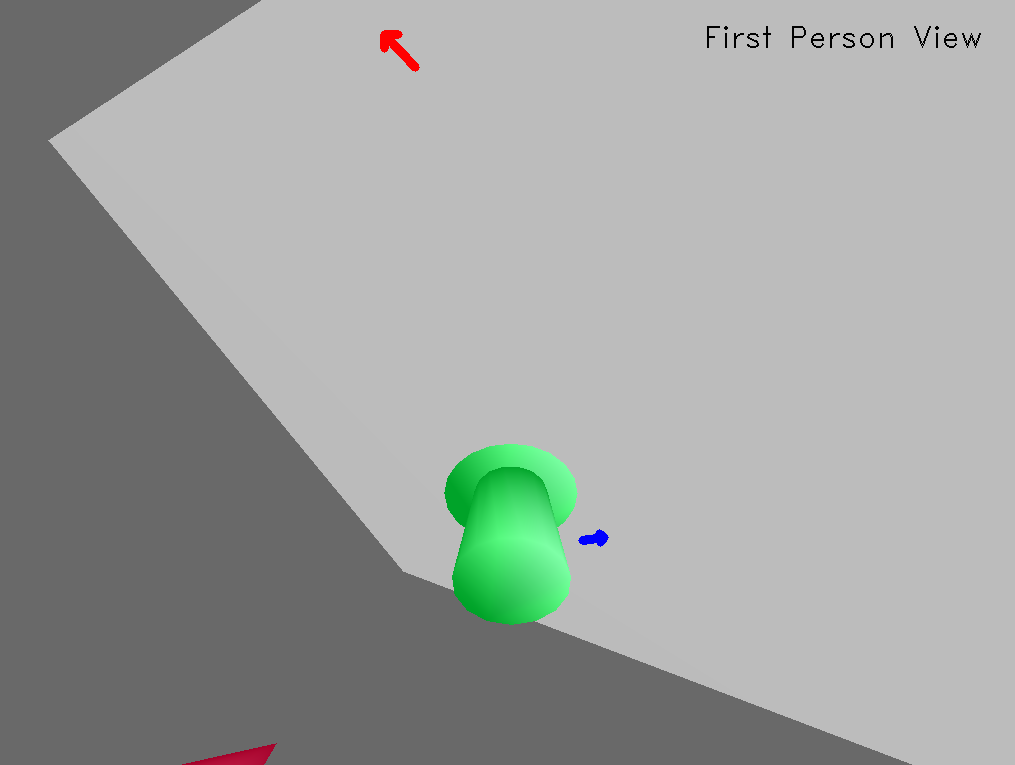}
   \end{minipage} &
   \begin{minipage}{2.9cm}
   \includegraphics[width=3cm]{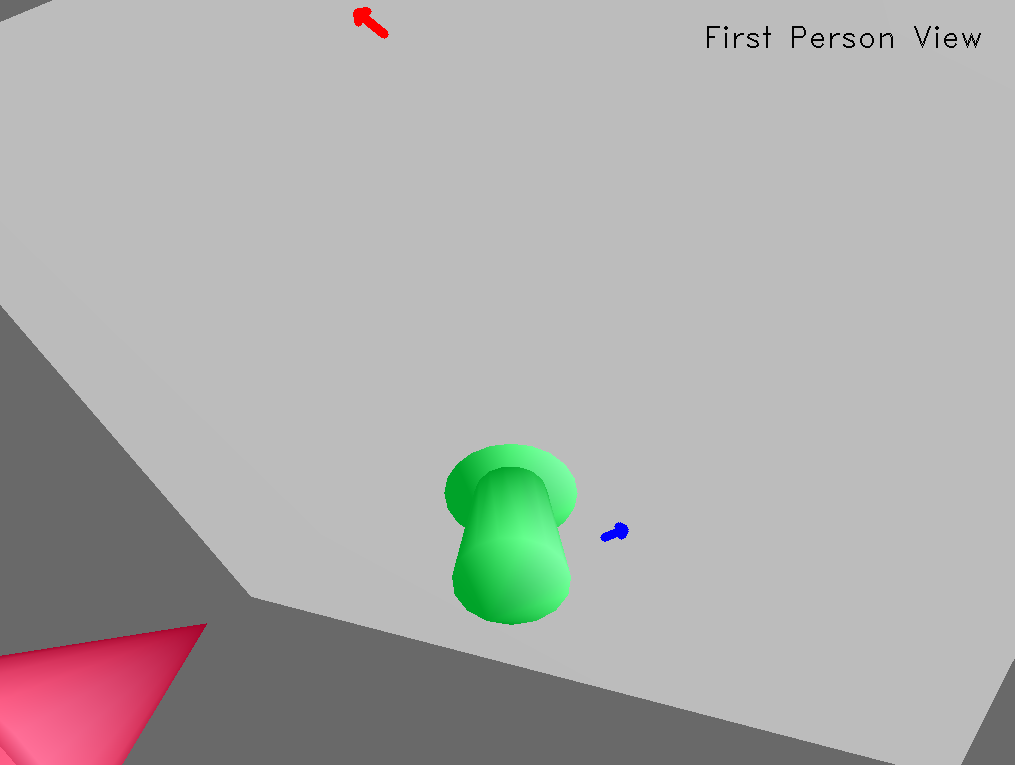}
   \end{minipage} &
   \begin{minipage}{2.9cm}
   \includegraphics[width=3cm]{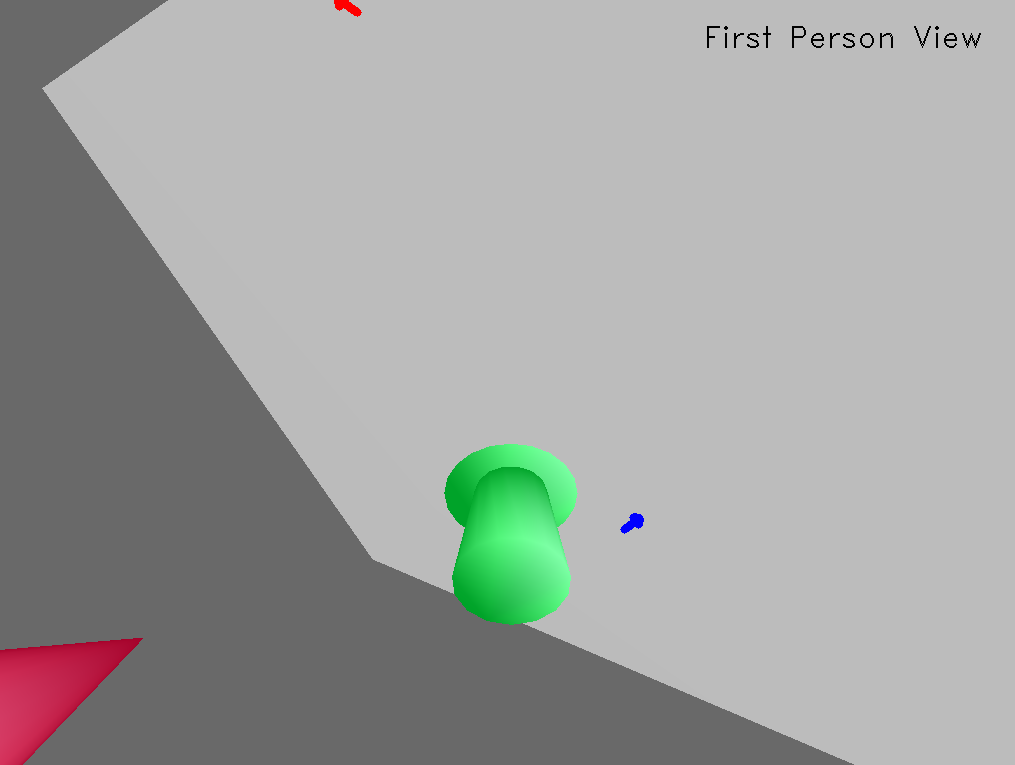}
   \end{minipage} &
   \begin{minipage}{2.9cm}
   \includegraphics[width=3cm]{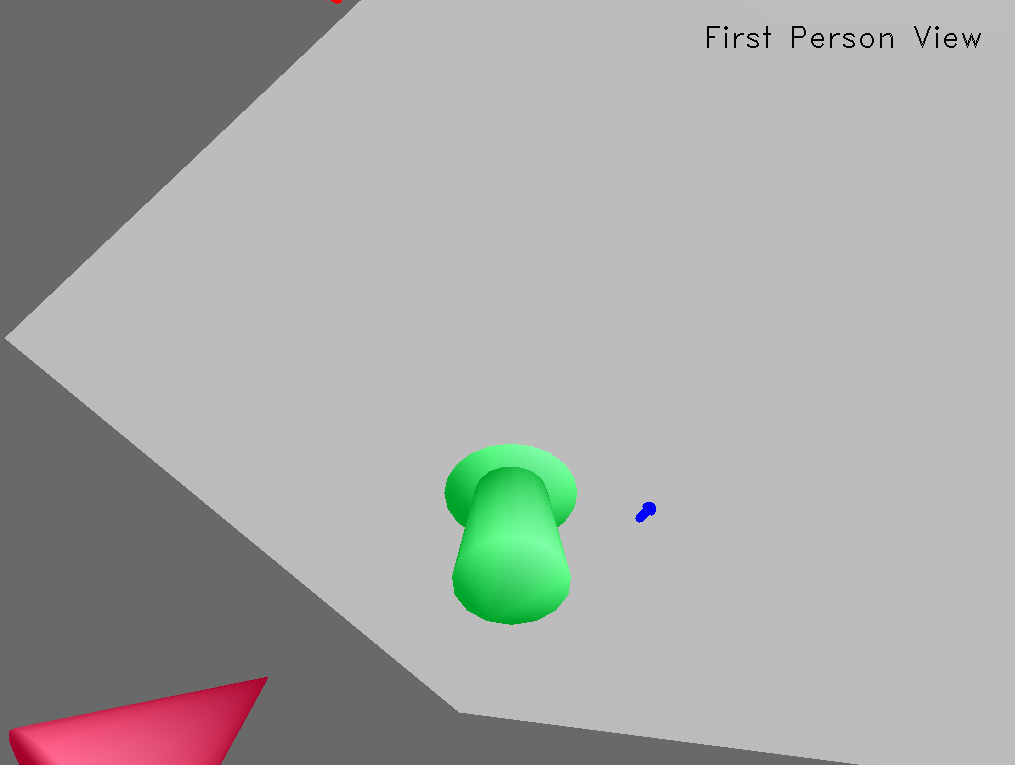}
   \end{minipage}\\
   \begin{minipage}{2.85cm}
   \includegraphics[width=3cm]{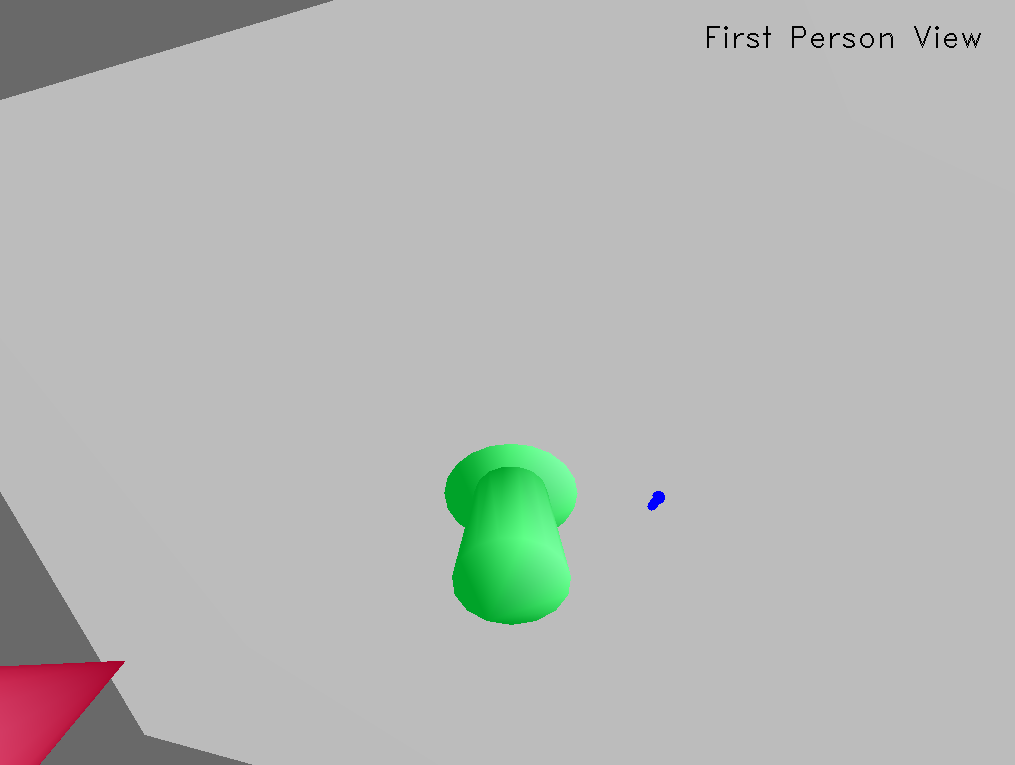}
   \end{minipage} &
   \begin{minipage}{2.9cm}
   \includegraphics[width=3cm]{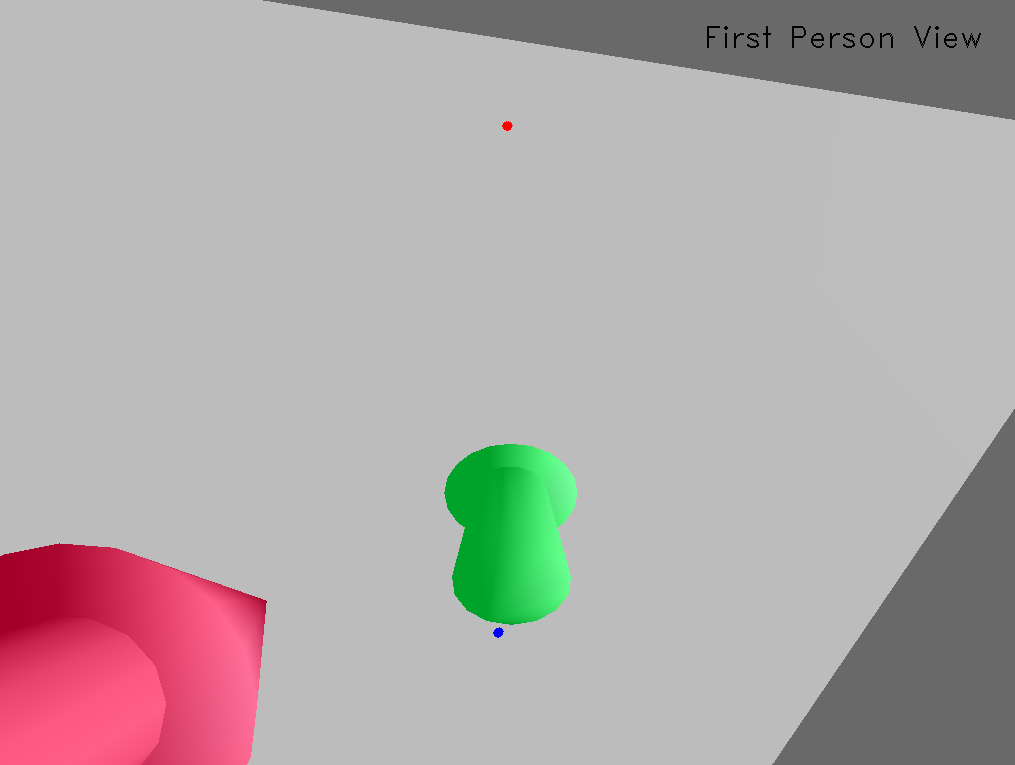}
   \end{minipage} &
   \begin{minipage}{2.9cm}
   \includegraphics[width=3cm]{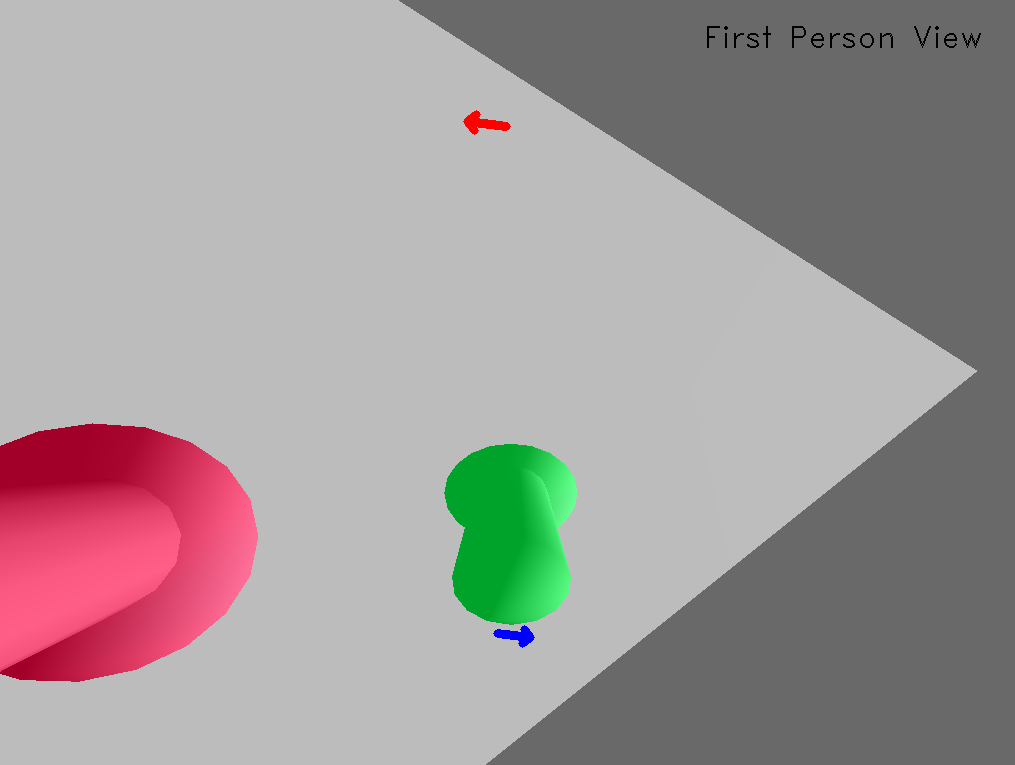}
   \end{minipage} &
   \begin{minipage}{2.9cm}
   \includegraphics[width=3cm]{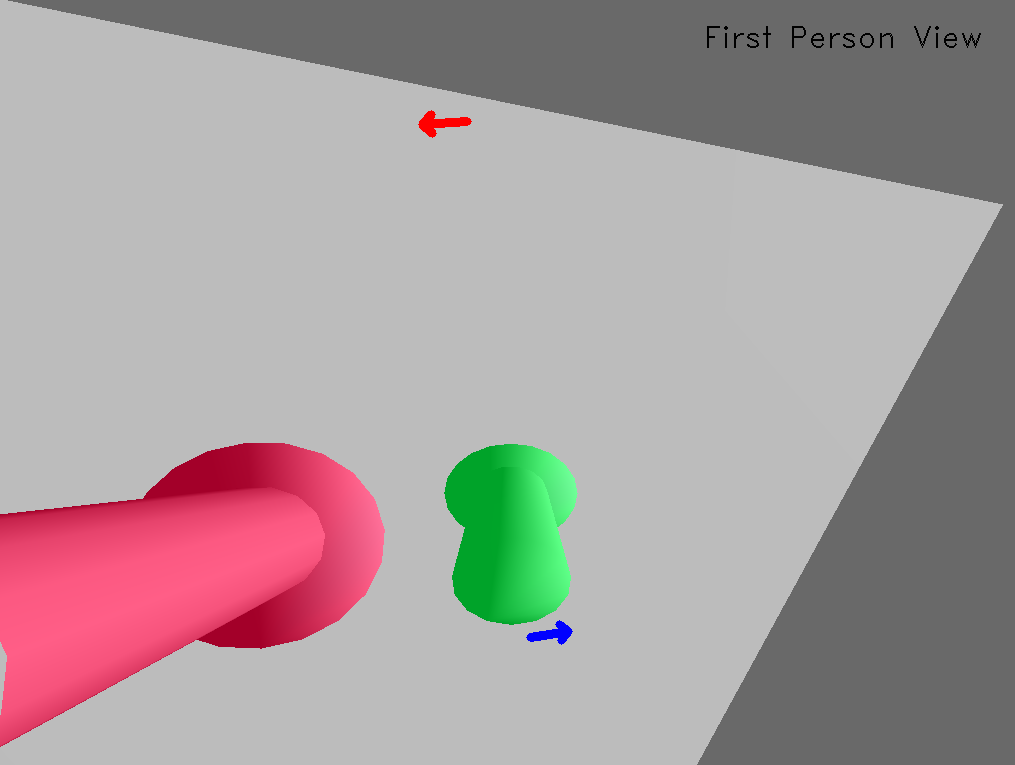}
   \end{minipage}\\
   \begin{minipage}{2.85cm}
   \includegraphics[width=3cm]{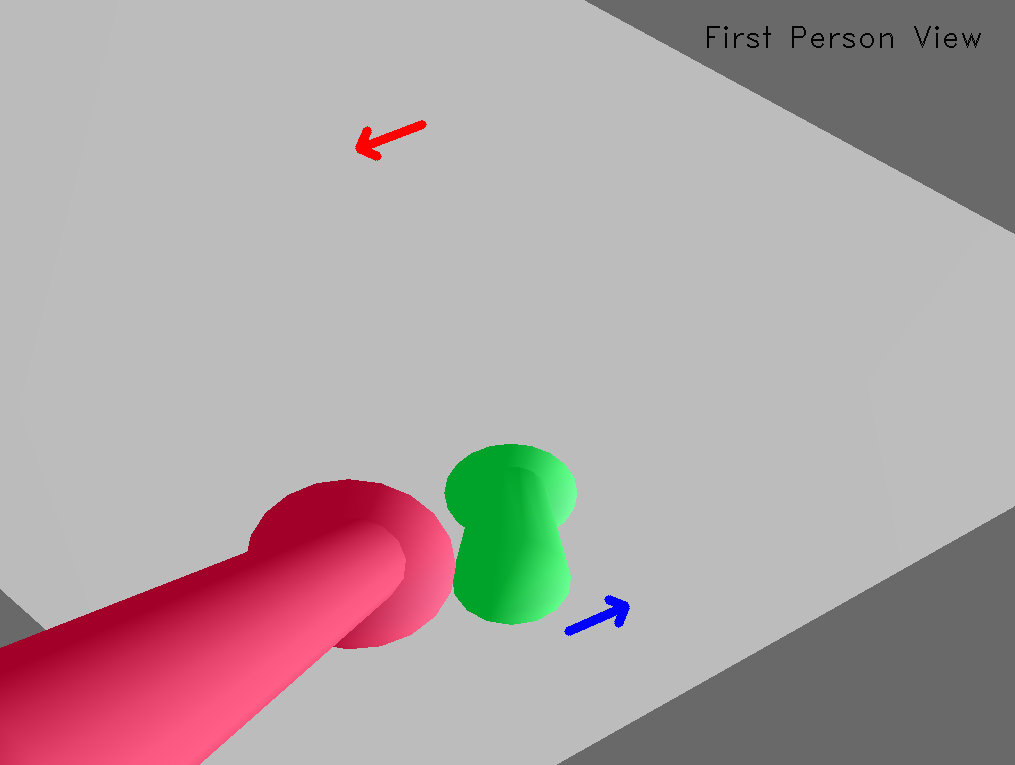}
   \end{minipage} &
   \begin{minipage}{2.9cm}
   \includegraphics[width=3cm]{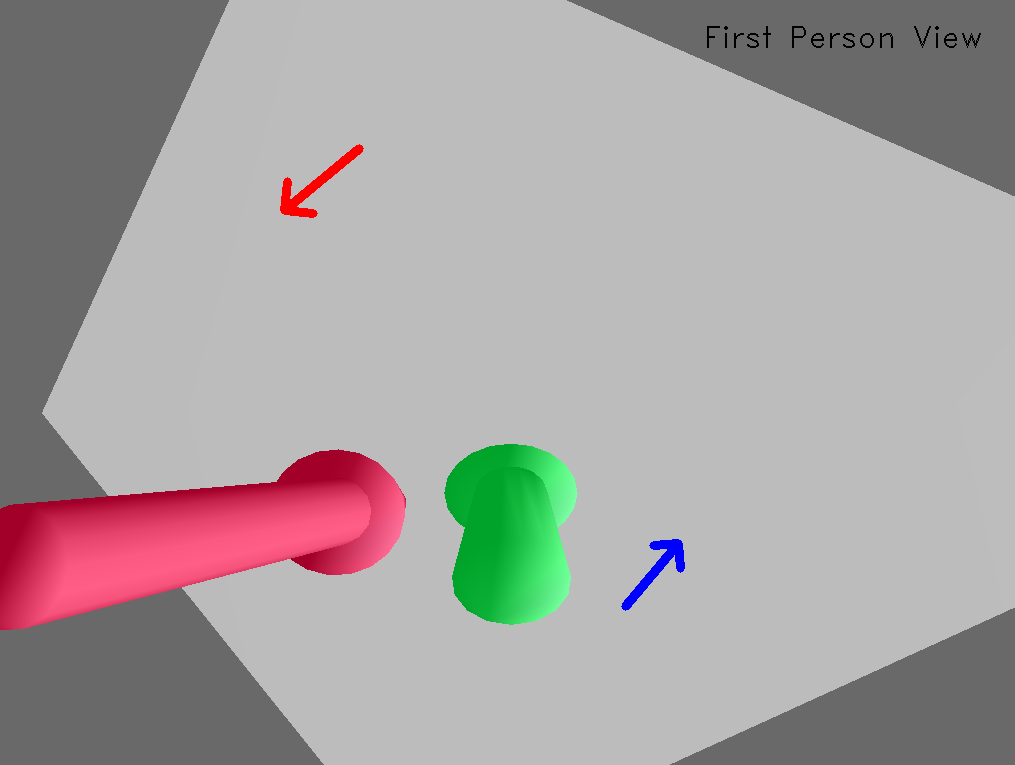}
   \end{minipage} &
   \begin{minipage}{2.9cm}
   \includegraphics[width=3cm]{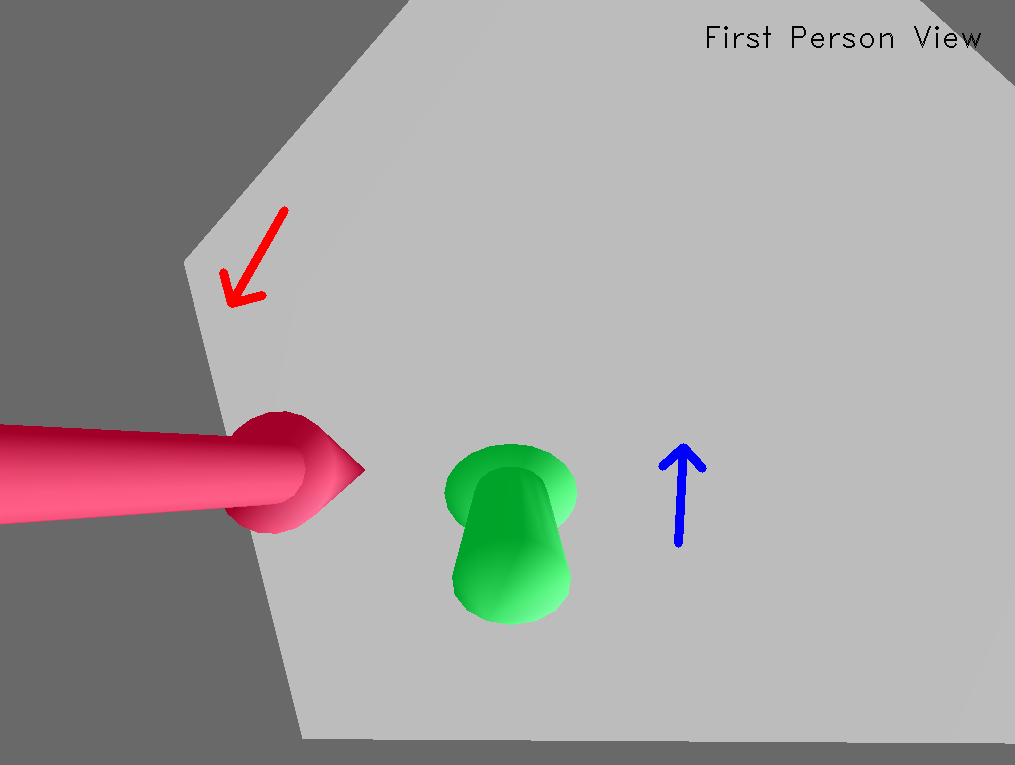}
   \end{minipage} &
   \begin{minipage}{2.9cm}
   \includegraphics[width=3cm]{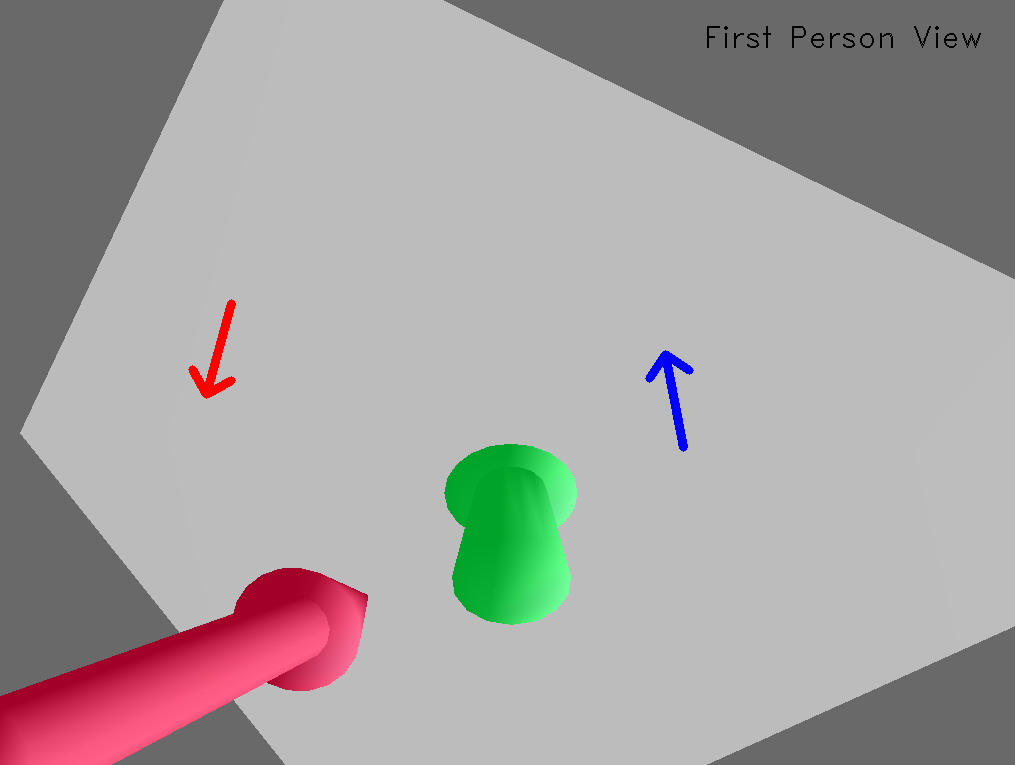}
   \end{minipage}\\
   \begin{minipage}{2.85cm}
   \includegraphics[width=3cm]{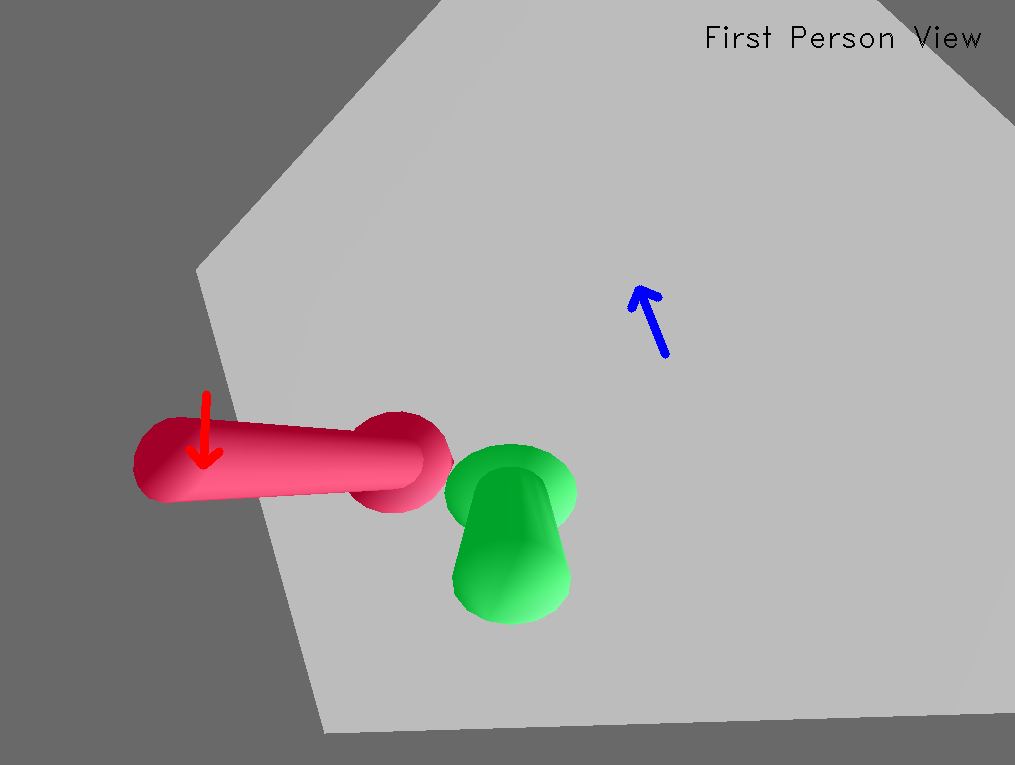}
   \end{minipage} &
   \begin{minipage}{2.9cm}
   \includegraphics[width=3cm]{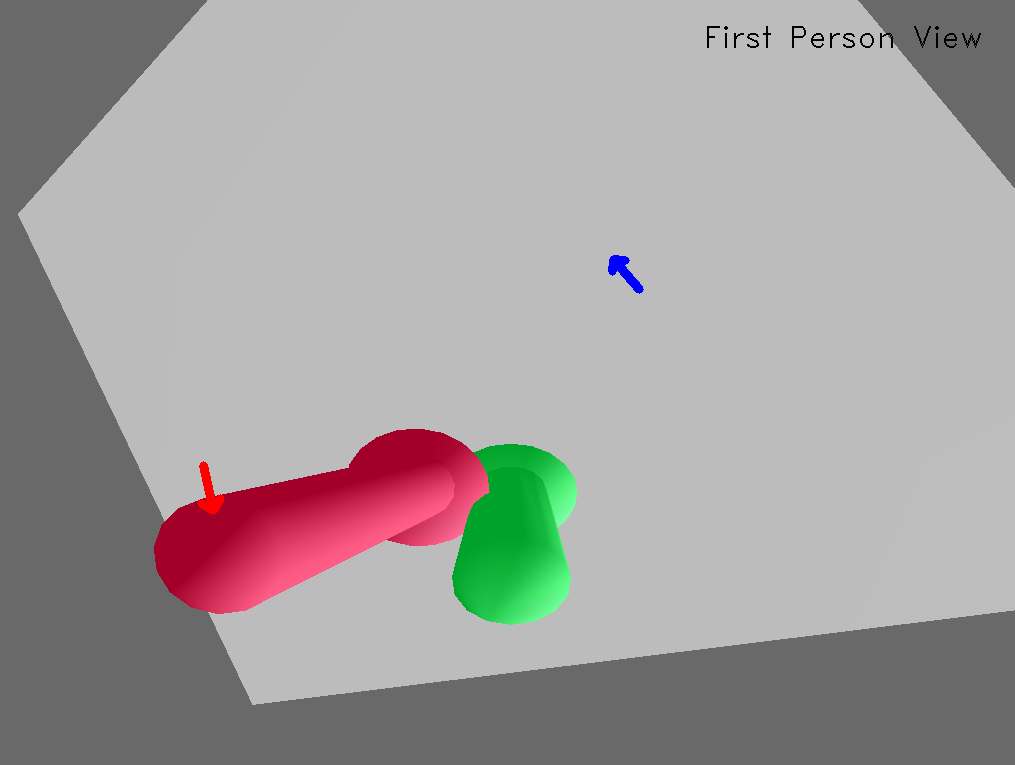}
   \end{minipage} &
   \begin{minipage}{2.9cm}
   \includegraphics[width=3cm]{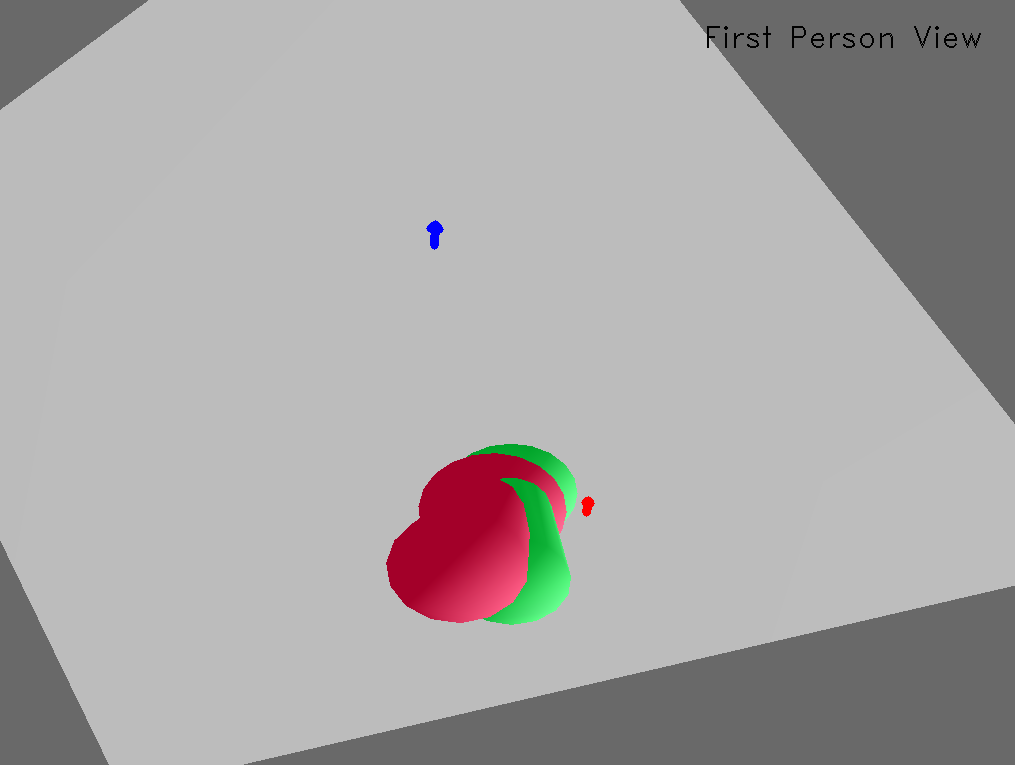}
   \end{minipage} &
   \begin{minipage}{2.9cm}
   \end{minipage}
   \end{tabular}
   \caption{\label{fig:biggerRollout3D}A full rollout example on the 3D environment, succeeded in 19 steps. Only the user view is displayed. Finger trajectories performed by the user agent are shown as blue and red arrows on the screen.}
   \end{center}
\end{figure}

\end{document}